%% file: main.tex
\documentclass{article}

\PassOptionsToPackage{square,sort,comma,numbers}{natbib}

\usepackage[final]{neurips_2024}

\usepackage[utf8]{inputenc} 
\usepackage[T1]{fontenc}    
\usepackage{hyperref}       
\usepackage{url}            
\usepackage{booktabs}       
\usepackage{amsfonts}       
\usepackage{microtype}      
\usepackage{xcolor}         
\usepackage{algorithm}
\usepackage{algpseudocode}

\usepackage{microtype}
\usepackage{graphicx}
\usepackage{subfigure}
\usepackage{booktabs} 
\usepackage{comment}
\usepackage{placeins}

\usepackage{hyperref}


\input{math_commands.sty}
\usepackage{algpseudocode}

\usepackage{amsmath}
\usepackage{amssymb}
\usepackage{mathtools}
\usepackage{amsthm}

\usepackage[capitalize]{cleveref}
\definecolor{cvprblue}{rgb}{0.21,0.49,0.74}
\usepackage{colortbl}

\theoremstyle{plain}

\title{Occam's Razor for Self Supervised Learning:\\What is Sufficient to Learn Good Representations?}

\author{%
  Mark Ibrahim \\
  FAIR, META\\
  \texttt{marksibrahim@meta.com} \\
  \And
  David Klindt \\
  Cold Spring Harbor Laboratory \\
  \texttt{klindt@cshl.edu} \\
  \And
  Randall Balestriero \\
  Brown University \\
  Computer Science Department \\
  \texttt{rbalestr@brown.edu}
}

\begin{document}

\maketitle

\begin{abstract}
\input{content/abstract}
\end{abstract}

\input{content/introduction}

\input{content/background}

\section{DIET: A Simplified Self Supervised Loss}

We first present in \cref{sec:DIET} the proposed DIET which is built by starting from a state-of-the-art SSL pipeline and simplifying it as much as possible. Thorough empirical validations on natural and medical images are provided in \cref{sec:SOTA,sec:medical} where we will see that many of the current SSL pipeline design are not necessary to learn good representations.

\vspace{-0.2cm}
\subsection{Simplifying Current Self Supervised Pipelines}
\label{sec:DIET}

The goal of this section is to introduce the proposed objective that we will use to contrast with current SSL objectives.


{\bf Simplification 1: from relative to absolute loss.} 
It has been brought to lights many times that the numerous variations of Self Supervised Learning take the form of comparing the inter-sample representations, aiming to {\em collapse} together the positive pairs, while ensuring that the entire representation does not collapse \cite{haochen2021provable,balestriero2022contrastive,cabannes2023active}. Within that formulation, SSL treats each sample as its own class that should have tightly knit representations, while being far away from all other samples, i.e., classes. We thus propose to replace the relative inter-sample objective with a cross-entropy loss using as target the index of the original datum--directly optimizing for that SSL is implicitly trying to solve. That is, given a dataset of $N$ samples $\{\vx_1,\dots,\vx_N\}$, define the class of sample $\vx_n$ with $n \in \{1,\dots,N\}$ to be $n$ itself.

{\bf Simplification 2: removal of the nonlinear projector.} We remove the usual nonlinear projector ($g_{\vgamma}$) and instead only use a liner classifier that maps the K-dimensional output of the originally considered model $f_{\vtheta}$ to the $N$ classes of the cross-entropy objective. We denote that linear classifier as ${\color{red}\mW \in \mathbb{R}^{N \times K}}$, 

{\bf Simplification 3: removal of teacher-student network and positive pairs.} Additionally, we remove the need to have positive pairs within each mini-batches, and also the possible presence of a teacher student network.

Leading to the final formulation
\begin{align}
    \mathcal{L}_{\rm\small DIET}(\vx_n)={\rm XEnt}({\color{red}\mW}{\color{blue}f_{\vtheta}}(\vx_n),n),\label{eq:DIET}
\end{align}
given a sample $\vx_n \in \mathbb{R}^{D}$. Thus, DIET performs unsupervised learning through the default supervised scheme, meaning that any progress made in the latter can be directly ported to DIET.  We propose its pseudo-code in \cref{algo:DIET} as well as the code to obtain a data loader providing the user with the indices ($n$).

\begin{algorithm}[t!]
\begin{minipage}{0.56\linewidth}
\begin{lstlisting}[language=Python,escapechar=\%,numbers=none]
# take any preferred DNN e.g. resnet50
# see %\lsComment{\cref{algo:architectures}}% for other examples
f = torchvision.models.resnet50() # %\lsComment{\color{blue}$f_{\vtheta}$}%

# f comes with a classifier so we remove it
K = f.fc.in_features
f.fc = nn.Identity()

# define DIET's linear classifier and XEnt
W = nn.Linear(K, N, bias=False) # %\lsComment{\color{red}$\mW$ in \cref{eq:DIET}}%
XEnt = nn.CrossEntropyLoss(label_smoothing=0.8)

# define dataset and train %\lsComment{(Fig.~\ref{fig:DIET})}%
train_dataset = DatasetWithIndices(train_dataset)  
train_loader = DataLoader(train_dataset, ...)

for x, n in train_loader:
    loss = XEnt(W(f(x)), n) # %\lsComment{\cref{eq:DIET}}%
    # backprop/optimizer/scheduler
\end{lstlisting}
\end{minipage}
\begin{minipage}{0.42\linewidth}
\begin{lstlisting}[language=Python,escapechar=\%,numbers=none]
from torch.utils.data import Dataset, DataLoader
from torchvision.datasets import CIFAR100

class DatasetWithIndices(Dataset):
    def __init__(self, dataset):
        self.dataset = dataset
    def __getitem__(self, n):
        # disregard the labels
        x, _ = self.dataset[n]
        return x, n
    def __len__(self):
        return len(self.dataset)

# example with CIFAR100
C100 = CIFAR100(root)
C100_w_ind = DatasetWithIndices(C100)
\end{lstlisting}
\end{minipage}
\label{algo:DIET}
\caption{\small \ul{DIET's algorithm and dataset loader}.}
\end{algorithm}

\subsection{Benefits for Practical Deployment and Theoretical Research}

There are many direct benefit of the DIET's objective emerging from its simplicity. We highlight both a theoretical and a practical benefit.

{\bf Benefit for theoretical research and provable guarantees.}~First, DIET opens numerous avenues for theoretical research. This is in sharp contrast with the original SSL methods. In fact, current SSL lacks of theoretical guarantees as all existing studies have derived optimality conditions at the projector's output \cite{wang2020understanding,tian2020makes,jing2021understanding,huang2021towards,haochen2021provable,dubois2022improving,zhang2021understanding,wang2021understanding} which is not the output of interest since the projector is thrown away after SSL training and the DNN's output and the projector's output greatly differ \cite{chen2020simple,chen2020improved,cosentino2022toward,bordes2022guillotine}. As a further demonstration of DIET's theory-friendliness, we propose in \cref{sec:linear_model} a theoretical study of DIET with a linear model $f_{\vtheta}$, in which case we are able to prove that DIET performs a low-rank decomposition of the input data matrix and provably recovers the data's principal components. Again, that last result highlights how \cref{eq:DIET} greatly reduces the barrier to derive novel theoretical resutls and guarantee for SSL.

\begin{figure}[t!]
    \centering
    \begin{minipage}{0.32\linewidth}
    \includegraphics[width=\linewidth]{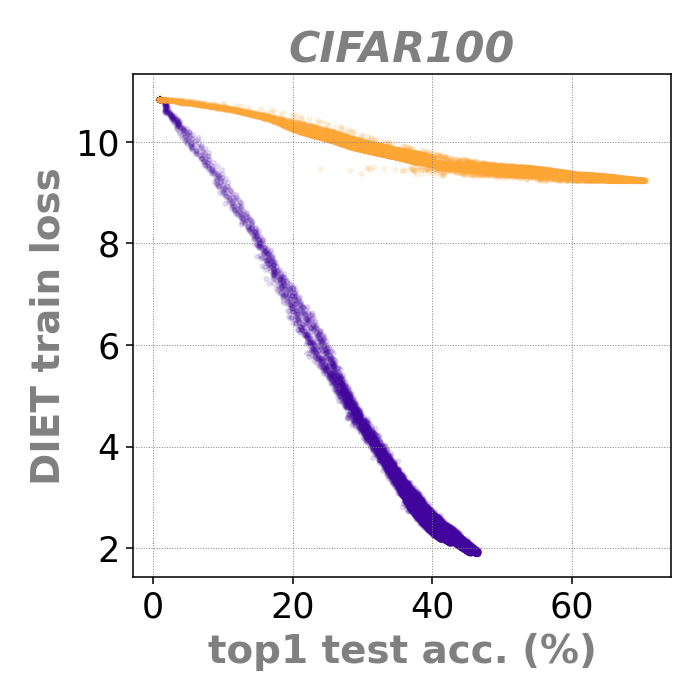}
    \end{minipage}
    \begin{minipage}{0.32\linewidth}
    \includegraphics[width=\linewidth]{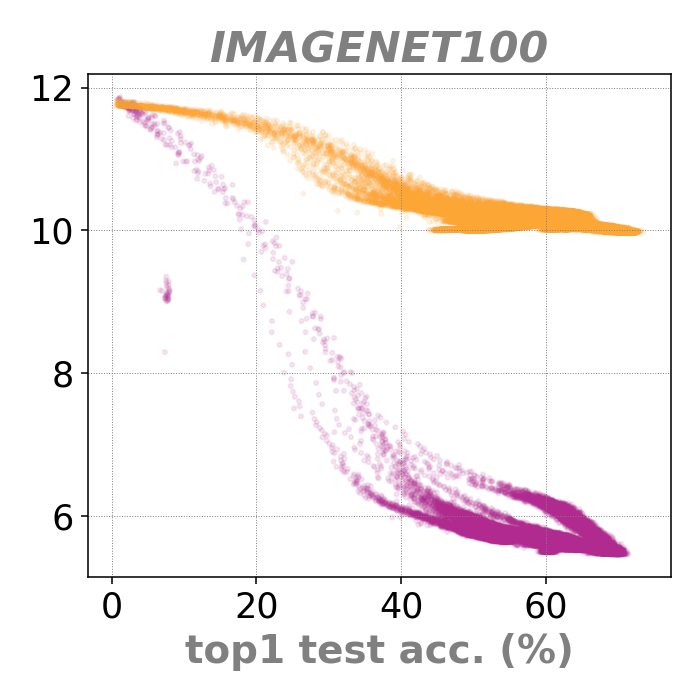}
    \end{minipage}
    \begin{minipage}{0.32\linewidth}
    \includegraphics[width=\linewidth]{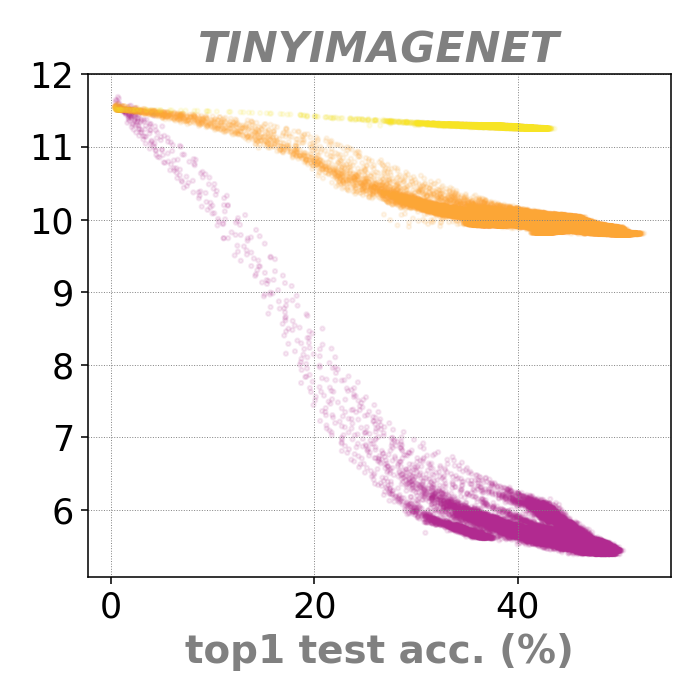}
    \end{minipage}
    \begin{minipage}{\linewidth}
    \caption{\small {\bf DIET's training loss is indicative of downstream test performance.}~We depict DIET's training loss ({\bf y-axis}) against the online test linear probe accuracy ({\bf x-axis}) for all the models, hyper-parameters, and training epochs. Yellow to purple correspond to different label smoothing which plays a role in DIET's convergence speed (\cref{sec:ablation}). For a given label smoothing parameter, there exists a strong relationship between \textbf{DIET}'s training loss and the downstream test accuracy enabling label-free quantitative quality assessment one's model. }
    \label{fig:accus}
    \end{minipage}
\end{figure}

\textbf{Benefits for practical development and deployment}
Second, the amount of code refactoring is minimal (recall \cref{algo:DIET}): there is no change required for the data loading pipelines as opposed to SSL which requires positive pairs, no need to specify teacher-student architectures, and no need to design a projector/predictor DNN. Second, DIET's implementation is not architecture specific as we validate on Resne(x)ts, ConvNe(x)ts, Vision Transformers and their variants. Furthermore, DIET does not introduce any additional hyper-parameters in addition to the ones already present in supervised learning--and because DIET's training loss is informative of test classification performances (\cref{fig:accus})--it opens the door to truely label-free SSL.

\subsection{A Simple Strategy Without the Bells of Whistles of SSL}

Despite DIET's simplicity, we could not find an existing method that considered it perhaps due to the common belief that dealing with hundreds of thousands of classes ($N$ in \cref{fig:DIET}, the training set size) would not produce successful training. As such, the closest method to ours is \textit{Exemplar CNN} \cite{alexey2015discriminative} which extracts a few patches from a given image dataset, and treats each of them as their own class; this way the number of classes is the number of extracted patches, which is made independent from $N$. A more recent method, \textit{Instance Discrimination} \cite{wu2018unsupervised} extends this by introducing inter-sample discrimination. However, they do so using a non-parametric softmax, {\em i.e.}, by defining a learnable bank of centroids to cluster training samples; for successful training those centroids must be regularized to prevent representation collapse. As we will compare in \cref{tab:C100}, DIET outperforms Instance Discrimination and Exemplar CNN while being simpler. Lastly, methods such as Noise as Targets \cite{bojanowski2017unsupervised} and DeepCluster \cite{caron2018deep} are quite far from DIET as (i) they perform clustering and use the datum's cluster as its class, {\em i.e.}, greatly reducing the dependency on $N$; and (ii) they perform clustering in the output space of the model {\color{blue} $f_{\vtheta}$} being learned which brings multiple collapsed solutions that force those methods to employ complicated mechanisms to ensure training to learn non-trivial representations. We note that while the added complexity enables those methods to scale to large datasets, it also greatly increases the performance sensitivity to the training hyper-parameters.

\section{A Simpler Loss Maintains Performances and Removes The Need For Cross-Validation}

To support the different claims we have made in the previous section, we will first explore natural image datasets in \cref{sec:SOTA} that including CIFAR100, Imagenet100, TinyImagenet, but also other datasets such as Food101 which have been challenging for SSL. We will then move to medical images in \cref{sec:medical} which are outside of the domain that SSL has been extensively optimized for. We will see there that DIET greatly outperforms those benchmarks without requiring any tuning or cross-validation. After having validated the ability of DIET to compete and often outperform SSL methods, we will spend \cref{sec:ablation} to probe the few hyper-parameters that govern DIET, in our case the label smoothing of the \texttt{XEnt} loss, and the training time. We will see that without label smoothing, DIET is often as slow as SSL methods to converge, and sometimes slower--but that high values of label smoothing greatly speed up convergence. 

Throughout our empirical validation, we will rigorously follow the experimental setup described in \cref{fig:hparams}. Our goal in adopting the same setup across experiments is to highlight the stability of DIET to dataset and architectural changes; careful tuning of those design choices should naturally lead to greater performance if desired.

\vspace{-0.2cm}
\subsection{DIET's simple objective is on par with SOTA on Natural Images}
\label{sec:SOTA}

We start the empirical validation of DIET on CIFAR100; following that, we will consider other common medium scale datasets, {\em e.g.}, TinyImagenet, and in particular we will consider datasets such as Food101, Flowers102 for which current SSL does not provide working solutions and for which the common strategy consists in transfer learning. We will see in those cases that applying DIET as-is on each dataset results in high-quality representations across different DNN architectures.

\begin{table}[t!]
    \centering
    \setlength{\tabcolsep}{0.32em}
    \renewcommand{\arraystretch}{0.6}
    \begin{minipage}{0.66\linewidth}
    \begin{tabular}{@{}c@{}|@{}c@{}}
    \begin{tabularx}{0.5\linewidth}{>{\hsize=1.5\hsize}X | >{\hsize=.5\hsize}X}
        \multicolumn{2}{c}{\em\cellcolor{blue!45}Resnet18}\\
        MoCoV2& 53.28$^\ast$\\
        SimSiam& 53.66$^\bullet$\\
        SimCLR&53.79$^\dagger$\\ 
        SimMoCo& 54.11$^\ast$\\
        ReSSL&54.66$^\bullet$\\
        SimCLR+adv & 55.51$^\dagger$\\
        MoCo& 56.10$^\ddagger$\\
        SimCLR&56.30$^\star$\\
        MoCo+CC& 57.65$^\ddagger$\\
        SimCLR&57.81$^\triangleright$\\
        DINO& 58.12$^\bullet$\\
        SimCO& 58.35$^\ast$\\
        SimCLR+DCL & 58.50$^\dagger$\\
        SimCLR&60.30$^\ddagger$\\
        SimCLR&60.45$^\bullet$\\
        W-MSE& 61.33$^\diamond$\\
        SimCLR+CC&61.91$^\ddagger$\\
        BYOL& 62.01$^\bullet$\\
        MoCoV2& 62.34$^\bullet$\\
        BYOL& 63.75$^\ddagger$\\
        \cellcolor{red!15}DIET & \cellcolor{red!15}63.77\\
        BYOL+CC& 64.62$^\ddagger$\\
        SimSiam& 64.79$^\ddagger$\\
        SwAV& 64.88$^\diamond$\\
        SimCLR&65.78$^\diamond$\\
        SimSiam+CC& 65.82$^\ddagger$\\ 
    \end{tabularx}
    &
    \begin{tabularx}{0.5\linewidth}{>{\hsize=1.5\hsize}X | >{\hsize=.5\hsize}X}
        \multicolumn{2}{c}{\em\cellcolor{blue!45}Resnet50}\\
        SimCLR& 52.04$^\dagger$\\
        MoCoV2& 53.44$^\ast$\\
        SimMoCo& 54.64$^\ast$\\
        SimCLR+adv& 57.71$^\dagger$\\
        SimCO& 58.48$^\ast$\\
        SimCLR& 61.10$^\star$\\
        SimCLR+DCL& 62.20$^\star$\\
        MoCoV3&69.00$^\triangleleft$\\
        \cellcolor{red!15}DIET&\cellcolor{red!15}69.91\\
        \multicolumn{2}{c}{\em\cellcolor{blue!45}Resnet101}\\
        SimCLR& 52.28$^\dagger$\\
        SimCLR+adv& 59.02$^\dagger$\\
        MoCoV3 &68.50$^\triangleleft$\\
        \cellcolor{red!15}DIET&\cellcolor{red!15}71.39\\
        \multicolumn{2}{c}{\em\cellcolor{blue!45}AlexNet}\\
        SplitBrain &39.00$^\Box$\\
        InstDisc &39.40$^\Box$\\
        DeepCluster& 41.90$^\Box$\\
        AND &47.90$^\Box$\\
        \cellcolor{red!15}DIET&\cellcolor{red!15}48.25\\
        SeLa&57.40$^\Box$\\
    \end{tabularx}
    \end{tabular}
    \end{minipage}
    \begin{minipage}{0.32\linewidth}
        \caption{\small {\bf DIET often outperforms benchmarks on CIFAR100.} We employ the settings of \cref{fig:hparams}, notice the consistent progression of the performance through architectures which is not easily achieved with standard SSL methods without per-architecture cross-validation.
    Benchmarks taken from
    $\dagger:$\cite{ho2020contrastive};
    $\ddagger:$\cite{peng2022crafting};
    $\ast$:\cite{zhang2022dual};
    $\bullet$:\cite{pham2022pros};
    $\diamond$:\cite{da2022solo};
    $\star$:\cite{yeh2022decoupled};
    $\triangleleft$:\cite{ren2022simple};
    $\triangleright$:\cite{yang2022identity};
    $\Box$:\cite{huang2022self}.
    }
    \label{tab:C100}
    \end{minipage}
    \vspace{-0.5cm}
\end{table}

\begin{table}[t!]
    \setlength{\tabcolsep}{0.32em}
    \renewcommand{\arraystretch}{0.8}
    \newcolumntype{P}[1]{>{\centering\arraybackslash}p{#1}}
    \centering
    \caption{\small {\bf DIET is competitive and works out-of-the-box across architectures.} We keep the settings of \cref{fig:hparams}, as per \cref{tab:C100}.
    Benchmarks from $1$:\cite{dubois2022improving}, $2:$\cite{ozsoy2022self}}
    \label{tab:tiny}
    \begin{minipage}{0.44\linewidth}
    \begin{tabular}{@{}c@{}|@{}c@{}}
    \multicolumn{2}{c}{\bf\ul{TinyImagenet}}\\
        \begin{tabularx}{0.5\linewidth}{X | X}\hline
         \multicolumn{2}{c}{\em\cellcolor{blue!45}Resnet18}\\
         SimSiam  & 44.54 $^\ddagger$\\
         SimCLR   & 46.21$^\ddagger$ \\
         BYOL     & 47.23$^\ddagger$ \\
         MoCo     & 47.98 $^\ddagger$\\
         SimCLR   & 48.70 $^1$\\
         DINO     & 49.20 $^1$\\
        \end{tabularx}&
        \begin{tabularx}{0.5\linewidth}{X | X}\hline
         \multicolumn{2}{c}{\em\cellcolor{blue!45}Resnet50}\\
            SimCLR &48.12 $^2$\\
            SimSiam &46.76 $^2$\\
            Spectral &49.86 $^2$\\
            CorInfoMax &54.86 $^2$\\
        \end{tabularx}\\ \hline
        \multicolumn{2}{c}{\cellcolor{red!15}DIET}\\
        \begin{tabularx}{0.5\linewidth}{>{\hsize=1.5\hsize}X | >{\hsize=.5\hsize}X}
        \em\cellcolor{blue!45} resnet18 & \cellcolor{red!15} 45.07 \\
        \em\cellcolor{blue!45} resnet34 & \cellcolor{red!15} 47.04 \\
        \em\cellcolor{blue!45} resnet101 & \cellcolor{red!15} 51.86 \\
        \em\cellcolor{blue!45} wide\_resnet50\_2 & \cellcolor{red!15} 50.03 \\
        \em\cellcolor{blue!45} resnext50\_32x4d & \cellcolor{red!15} 52.45 \\
        \em\cellcolor{blue!45} densenet121 & \cellcolor{red!15} 49.38 \\
        \end{tabularx}&
        \begin{tabularx}{0.5\linewidth}{>{\hsize=1.5\hsize}X | >{\hsize=.5\hsize}X}
            \em\cellcolor{blue!45} resnet50 & \cellcolor{red!15} 51.66 \\
            \em\cellcolor{blue!45} convnext\_tiny & \cellcolor{red!15} 50.88 \\
            \em\cellcolor{blue!45} convnext\_small & \cellcolor{red!15} 50.05 \\
            \em\cellcolor{blue!45} MLPMixer & \cellcolor{red!15} 39.32 \\
            \em\cellcolor{blue!45} swin\_t & \cellcolor{red!15} 50.80 \\
            \em\cellcolor{blue!45} vit\_b\_16 & \cellcolor{red!15} 48.38 \\
        \end{tabularx}\\
    \end{tabular}
    \end{minipage}
    \begin{minipage}{0.55\linewidth}
    \begin{tabular}{@{}c@{}|@{}c@{}}
    \multicolumn{2}{c}{\bf\ul{Imagenet-100 (IN100)}}\\
    \begin{tabularx}{0.5\linewidth}{X | X}\hline
            \multicolumn{2}{c}{\em\cellcolor{blue!45}Resnet18}\\
            SimMoCo &58.20$^\ast$\\
            MocoV2  &60.52$^\ast$\\
            SimCo   &61.28 $^\ast$\\
            W-MSE2  &69.06 $^2$\\
            ReSSL   &74.02$^\bullet$\\
            DINO    &74.16$^\bullet$\\
            MoCoV2  &76.48$^\bullet$\\ 
            BYOL    &76.60$^\bullet$\\
            SimCLR  &77.04$^2$\\
            SimCLR  &78.72$^2$\\
            MocoV2  &79.28$^2$\\
            VICReg  &79.40$^2$\\
            BarlowTwins  &80.38$^2$\\
        \end{tabularx}&
        \begin{tabularx}{0.5\linewidth}{>{\hsize=1.4\hsize}X | >{\hsize=.6\hsize}X}\hline
            \multicolumn{2}{c}{\em\cellcolor{blue!45}Resnet50}\\
            MoCo+Hyper.     &75.60 $^\star$\\
            MoCo+DCL        &76.80 $^\star$\\
            MoCoV2 + Hyper. &77.70 $^\star$\\
            BYOL            &78.76 $^2$\\
            MoCoV2 + DCL    &80.50 $^\star$\\
            SimCLR          &80.70 $^\star$\\
            SimSiam         &81.60$^2$\\
            SimCLR + DCL    &83.10 $^\star$\\
        \end{tabularx}\\\hline
        \multicolumn{2}{c}{\cellcolor{red!15}DIET}\\
        \begin{tabularx}{0.5\linewidth}{>{\hsize=1.5\hsize}X | >{\hsize=.5\hsize}X}
        \em\cellcolor{blue!45} resnet18 & \cellcolor{red!15} 64.31 \\
        \em\cellcolor{blue!45} wide\_resnet50\_2 & \cellcolor{red!15} 71.92 \\
        \em\cellcolor{blue!45} resnext50\_32x4d & \cellcolor{red!15} 73.07 \\
        \em\cellcolor{blue!45} densenet121 & \cellcolor{red!15} 67.46 \\
            \em\cellcolor{blue!45} convnext\_tiny & \cellcolor{red!15} 69.77\\
        \end{tabularx}&
        \begin{tabularx}{0.5\linewidth}{>{\hsize=1.5\hsize}X | >{\hsize=.5\hsize}X}
            \em\cellcolor{blue!45} resnet50& \cellcolor{red!15} 73.50 \\
            \em\cellcolor{blue!45} convnext\_small & \cellcolor{red!15} 71.06 \\
            \em\cellcolor{blue!45} MLPMixer & \cellcolor{red!15} 56.46 \\
            \em\cellcolor{blue!45} swin\_t & \cellcolor{red!15} 67.02 \\
            \em\cellcolor{blue!45} vit\_b\_16 & \cellcolor{red!15} 62.63 \\
        \end{tabularx}
    \end{tabular}
    \end{minipage}
    \vspace{-0.5cm}
\end{table}

{\bf DIET achieves high performance on CIFAR100:}~Let's first consider CIFAR100 \cite{krizhevsky2009learning} with a few variations of Resnet \cite{he2016deep} and AlexNet \cite{krizhevsky2014one} architectures. To accommodate the $32 \times 32$ resolution, we follow the standard procedure to slightly modify the ResNet architecture: the first convolution layer sees its kernel size go from 7$\times$7 to 3 $\times$ 3 and its stride reduced from 2 to 1; the first max pooling layer is removed (details in \cref{algo:architectures}). On Alexnet, a few non-SSL baselines are available: SplitBrain \cite{zhang2017split}, DeepCluster \cite{caron2018deep}, InstDisc \cite{wu2018unsupervised}, AND \cite{huang2019unsupervised}, SeLa \cite{asano2019self}, and ReSSL \cite{zheng2021ressl}. The models are trained with the DIET objective ({\cref{eq:DIET}), and linear evaluation is employed to judge the quality of the learned representation on the original classification task.
We report results \cref{tab:C100} where we observe that DIET is able to match and often slightly exceed current SSL methods. In particular, even though CIFAR100 is a relatively small dataset, increasing the DNN capacity, {\em i.e.}, from Resnet18 to Resnet101 does not exhibit any overfitting using DIET (similar generalization benefits in Sec. \ref{sec:medical}).

{\bf DIET is competitive out-of-the-box across architectures on TinyImagenet and ImageNet100.}~We continue our empirical validation with the more challenging Imagenet100 (IN100) \cite{tian2020contrastive} dataset which consists of 100 classes of the full Imagenet-1k dataset, the list of classes can be found online\footnote{\url{https://github.com/HobbitLong/CMC/blob/master/imagenet100.txt}}, and the TinyImagenet \cite{le2015tiny} dataset which consists of $200$ classes with lower resolution images. We broaden the considered space of architectures to not only include the Resnet variants, but also SwinTransformers \cite{liu2021swin}, VisionTransforms \cite{dosovitskiy2020image}, Densenets \cite{huang2017densely}, ConvNexts \cite{liu2022convnet}, WideResnets \cite{zagoruyko2016wide}, ResNexts \cite{xie2017aggregated}, and the MLPMixer \cite{tolstikhin2021mlp}. We report those results in \cref{tab:tiny} where we observe that DIET is now around the average performance of the multiple SSL methods combined. As most SSL methods have been thoroughly tuned for Imagenet style tasks, we expect those benchmarks to be more challenging. That being said, the results from \cref{tab:tiny} demonstrate how DIET handles out-of-the-box any architecture change --even for different architecture families, e.g., with and without self-attention. As we will see in \cref{sec:medical}, when considering other data modality out of the scope of current SSL methods, DIET achieves state-of-the-art performances.

\begin{table*}[t!]
    \centering
    \renewcommand{\arraystretch}{0.8}
    \setlength{\tabcolsep}{0.1em}
    \caption{\small {\bf DIET trained on small datasets competes with Imagenet pre-trained SSL.} 
    We also report performances for a ViT based architecture (SwinTiny) to demonstrate the ability of DIET to handle different models out-of-the-box following \cref{fig:hparams}. Benchmarks from $\dagger$:\cite{yang2022identity}, +:\cite{ericsson2021well}}
    \begin{tabular}{l|l|l|l|l|c|c|c|c|c|c}
       \multirow{3}{*}{Arch.} &\multirow{3}{*}{Pretrain} & \multirow{3}{*}{Frozen} & &  {\em Aircraft}   & {\em DTD    }   & {\em Pets}      & {\em Flower}    & {\em CUB-200} & {\em Food101} &{\em Cars}\\ 
       &&&N=&6667&1880&2940&1020&11788&68175&6509\\
       &&&C=&100&47&37&102&200&101&196\\\hline\hline
        \multirow{4}{*}{\em Resnet18}&\multirow{3}{*}{IN100$^\dagger$}&Yes&SimCLR        & 24.19       &   54.35   &   46.46   & 75.00     &16.73 &-&-\\
        &&&+CLAE & 25.87       &   52.12   &   43.55   & 76.82     &17.58 &-&-\\
        &&&+IDAA & 26.02       &   54.97   &   46.76   & 77.99     &18.15 &-&-\\\cline{2-11}
        &None&No&DIET & \cellcolor{red!15}37.29 & \cellcolor{red!15}50.62 & \cellcolor{red!15}64.06 & \cellcolor{red!15}72.01 &\cellcolor{red!15} 33.03 &\cellcolor{red!15} 62.00 &\cellcolor{red!15} 42.55\\ \hline\hline
        \multirow{17}{*}{\em Resnet50}&\multirow{13}{*}{IN-1k$^+$}&\multirow{13}{*}{Yes}&InsDis  & 36.87 & 68.46 & 68.78 & 83.44 & - & 63.39 & 28.98\\
        &  &  &MoCo    & 35.55 & 68.83 & 69.84 & 82.10 & - & 62.10 & 27.99\\
        &  &  &PCL.    & 21.61 & 62.87 & 75.34 & 64.73 & - & 48.02 & 12.93\\
        &  &  &PIRL    & 37.08 & 68.99 & 71.36 & 83.60 & - & 64.65 & 28.72\\
        &  &  &PCLv2   & 37.03 & 70.59 & 82.79 & 85.34 & - & 64.88 & 30.51\\
        &  &  &SimCLR  & 44.90 & 74.20 & 83.33 & 90.87 & - & 67.47 & 43.73\\
        &  &  &MoCov2  & 41.79 & 73.88 & 83.30 & 90.07 & - & 68.95 & 39.31\\
        &  &  &SimCLRv2& 46.38 & 76.38 & 84.72 & 92.90 & - & 73.08 & 50.37\\
        &  &  &SeLav2  & 37.29 & 74.15 & 83.22 & 90.22 & - & 71.08 & 36.86\\
        &  &  &InfoMin & 38.58 & 74.73 & 86.24 & 87.18 & - & 69.53 & 41.01\\
        &  &  &BYOL    & 53.87 & 76.91 & 89.10 & 94.50 & - & 73.01 & 56.40\\
        &  &  &DeepClusterv2& 54.49 & 78.62 & 89.36 & 94.72 & - & 77.94 & 58.60\\
        &  &  &Swav    & 54.04 & 77.02 & 87.60 & 94.62 & - & 76.62 & 54.06\\ \cline{2-11}
        &None&No&DIET& \cellcolor{red!15}44.81 & \cellcolor{red!15}51.75 & \cellcolor{red!15}67.08 & \cellcolor{red!15}73.32 & \cellcolor{red!15}41.03 & \cellcolor{red!15}71.58 & \cellcolor{red!15}55.82\\ \cline{2-11}\hline
        \multirow{1}{*}{\em SwinTiny}&None&No&DIET& \cellcolor{red!15}33.15 & \cellcolor{red!15}51.88 & \cellcolor{red!15}58.06 & \cellcolor{red!15}70.78 & \cellcolor{red!15}32.11 & \cellcolor{red!15}68.86 &\cellcolor{red!15} 47.12\\\hline
        \multirow{1}{*}{\em Convnext-S}&None&No&DIET& \cellcolor{red!15}43.13 & \cellcolor{red!15}49.52 & \cellcolor{red!15}61.72 & \cellcolor{red!15}67.72 & \cellcolor{red!15}31.44 & \cellcolor{red!15}69.84 &\cellcolor{red!15} 40.63\\
    \end{tabular}
    \label{tab:transfer}
\end{table*}

{\bf DIET trained on small datasets competes with more complex Imagenet pre-trained SSL methods.}~We conclude the first part of our empirical validation by considering small datasets that are commonly handled by SSL through transfer learning: Aircraft \cite{maji13fine-grained}, DTD \cite{cimpoi14describing}, Pets \cite{parkhi2012cats}, Flowers \cite{nilsback2008automated}, CUB200 \cite{WahCUB_200_2011}, Food101 \cite{bossard14}, Cars \cite{KrauseStarkDengFei-Fei_3DRR2013}, where the numbers of training samples is much smaller than the standard Imagenet dataset, and where the image distribution are often much less diverse, e.g., focusing only on aircraft images. The current best solution to solve those tasks is to pretrain one's favorite SSL method on a larger dataset such as Imagenet100 or Imagenet-1k where SSL is known to be state-of-the-art, and to transfer the learned representation. By contrast, DIET finally provides an alternative approach by training directly on the considered small dataset. We report those results in \cref{tab:transfer} where we see that DIET competes with  or in some cases outperforms SSL models pretrained on much larger data. We hope that this will encourage more reliable in-distribution representation learning as opposed to transfer learning, which can be difficult to rely on when there is a domain shift between the pre-training and task image distributions \citep{zhuang2020comprehensive}.

We also find DIET can even outperform supervised learning methods in some cases when few data-labels are available. Furthermore, we show DIET's learning objective can be used with specialized network architectures such as scattering networks. We refer interested reader to \cref{app:diet_v_supervised} for details of these explorations.

\vspace{-0.2cm}
\subsection{DIET Provides State-of-the-Art Performance on Medical Images}
\label{sec:medical}

We now propose a more challenging comparison on medical images--an important modality that is often left behind when developing and tuning new SSL methods. We will see that DIET is able to produce state-of-the-art performances out-of-the-box.

We evaluate DIET training from scratch on three datasets from the MedMNISTv2 benchmark \cite{Yang_2023} (i) PathMNIST consisting of $90,000$ training images and $7,180$ test images, (ii) DermaMNIST consisting of $10,015$ training and $2,005$ test images, and finally (iii) BloodMNIST consisting of $17,092$ training and $3,421$ test images. To match a realistic unsupervised representation learning scenario, we employ for each method the hyper-parameters that work well on CIFAR100, and assume no labels are available for SSL training.
For DIET, we use the same hyperparameters used for CIFAR100. For the baseline SSL methods, we select a variety of methods including a contrastive method (SimCLR), a momentum based method (MoCov2), and a recent non-contrastive method (ViCReg). For those, we use the default hyperparameters from \cite{susmelj2020lightly} which yield good performance (> 80\%) on CIFAR10, a comparable small dataset consisting of $60,000$ images.

We find that although all algorithms achieve high training accuracy via a linear probe as shown in \cref{tab:medmnist}, the features learned by the baseline SSL methods do not generalize well to the test sets. By contrast, DIET achieves much higher performances (also see \cref{app:medmnist} for DIET with ViT). We also show training curves for both the DIET loss and the online training accuracy which exhibit stable convergence out-of-the-box with the same hyper-parameters used throughout the paper in \cref{fig:medmnist-diet-loss}. In addition, DIET's simplicity makes it faster to reach a given number of epochs, specifically for ResNet18, DIET is 1.75x faster than SimCLR (and 1.72x faster than VICReg), thanks to DIET's simple learning objective.

\begin{table}[t!]
  \begin{minipage}[c]{0.6\textwidth}
    \setlength{\tabcolsep}{0.4em}
    \renewcommand{\arraystretch}{0.8}
    \begin{tabular}{lrr|rr|rr}
        \toprule
        dataset & \multicolumn{2}{c|}{bloodmnist} & \multicolumn{2}{c|}{dermamnist} & \multicolumn{2}{c}{pathmnist} \\
         & train & test & train & test & train & test \\
        \midrule
        DIET & {\color{gray}87.16}&89.24 & {\color{gray}73.13}&73.92 & {\color{gray}87.05}&{\bf 44.53} \\
        DIET+ & {\color{gray}91.18}&{\bf 90.44} & {\color{gray}73.73}&{\bf 74.21} & {\color{gray}88.96}&{\bf 44.54} \\
        MoCov2 & {\color{gray}87.30}&53.70 & {\color{gray}70.99}&66.88 & {\color{gray}85.12}&18.97 \\
        SimCLR & {\color{gray}86.26}&14.56 & {\color{gray}69.23}&66.88 & {\color{gray}87.16}&11.80 \\
        VICReg & {\color{gray}88.78}&47.18 & {\color{gray}70.80}&66.78 & {\color{gray}87.60}&11.31 \\
        \midrule
        Transfer & {\color{gray}86.68}&88.13 & {\color{gray}73.58}&74.06 & {\color{gray}87.84}&59.37 \\
        \bottomrule
        \end{tabular}
  \end{minipage}\hfill
  \begin{minipage}[c]{0.4\textwidth}
    \caption{\small Performance on MedMNIST datasets using a Resnet18 (ViT provided in \cref{app-tab:medmnist-diet-vit}). DIET+ refers to the same DIET model
            trained for the same number of GPU hours as other models.
                VICReg is trained with the same hyperparameters 
                as SimCLR with SGD 6e-2. Transfer is pretrained on ImageNet and fixed with a linear probe.}
    
  \end{minipage}
  \label{tab:medmnist}
\end{table}

\vspace{-0.2cm}
\subsection{DIET's Dependency on Data-Augmentation, Training Time and Batch Size}
\label{sec:ablation}

The aim of this section is to better inform practitioners about the role of Data-Augmentations (DA), training time, and label smoothing in DIET's performances; as well as sensitivity to batch size, which is crucial for single-GPU training.


\begin{table}[t!]
    \centering
    \setlength{\tabcolsep}{0.2em}
    \caption{Ablation studies indicate that {\bf DIET benefits from longer training and stronger data augmentation while being robust to architecture and batch-size changes}. We report top1 test accuracy on CIFAR100 with varying training epochs ({\bf top left}), on TinyImagenet with varying DA pipelines (\cref{algo:DA}), and on TinyImagenet with 3k training epochs and with varying batch-size ({\bf bottom}) with learning rate $0.001\frac{\texttt{bs}}{256}$; additional comparisons on MedMNIST \cref{app:BS_CV_medmnist}.}
    \label{tab:epochs}\label{tab:BS_CV}\label{tab:DA}
    \renewcommand{\arraystretch}{0.8}
    \begin{minipage}{0.59\linewidth}
    \begin{tabular}{lrrrrrrr}
    \toprule
     Epochs    & 50    & 100    & 200    & 500    & 1000    & 5000    & 10000    \\ \midrule
     resnet18  & 33.46 &  42.94 &  48.24 &  54.54 &   58.81 &   62.63 &    63.29 \\
     resnet50  & 37.71 &  47.86 &  54.04 &  60.23 &   64.24 &   69.51 &    69.91 \\
     resnet101 & 34.03 &  46.59 &  54.3  &  60.8  &   64.71 &   70.56 &    71.39 \\
    \bottomrule
    \end{tabular}
    \end{minipage}
    \begin{minipage}{0.39\linewidth}
    \begin{tabular}{l|lll}
    \toprule
     DA strength & 1    & 2    & 3  \\
    resnet18& 31.48 & 43.62 & 43.88 \\
    resnet34& 32.93 & 45.60  & 45.75 \\
    resnet50& 40.24 & 48.80  & 50.81 \\
    resnet101& 40.07& 49.74 & 50.76 \\
    \bottomrule
    \end{tabular}
    \end{minipage}
    \begin{tabular}{lrrrrrrrr}\toprule
     batch-size&  8   & 16   & 32   & 64   & 128   & 256   & 512   & 1024   \\ 
     resnet18&32.9 & 37.9 & 42.7 & 43.4 &  43.3 &  43.7 &  43.7 &   42.6 \\ \bottomrule
    \end{tabular}
\end{table}

\begin{figure*}[t!]
    \centering
    \begin{minipage}{\linewidth}
    \includegraphics[width=\linewidth]{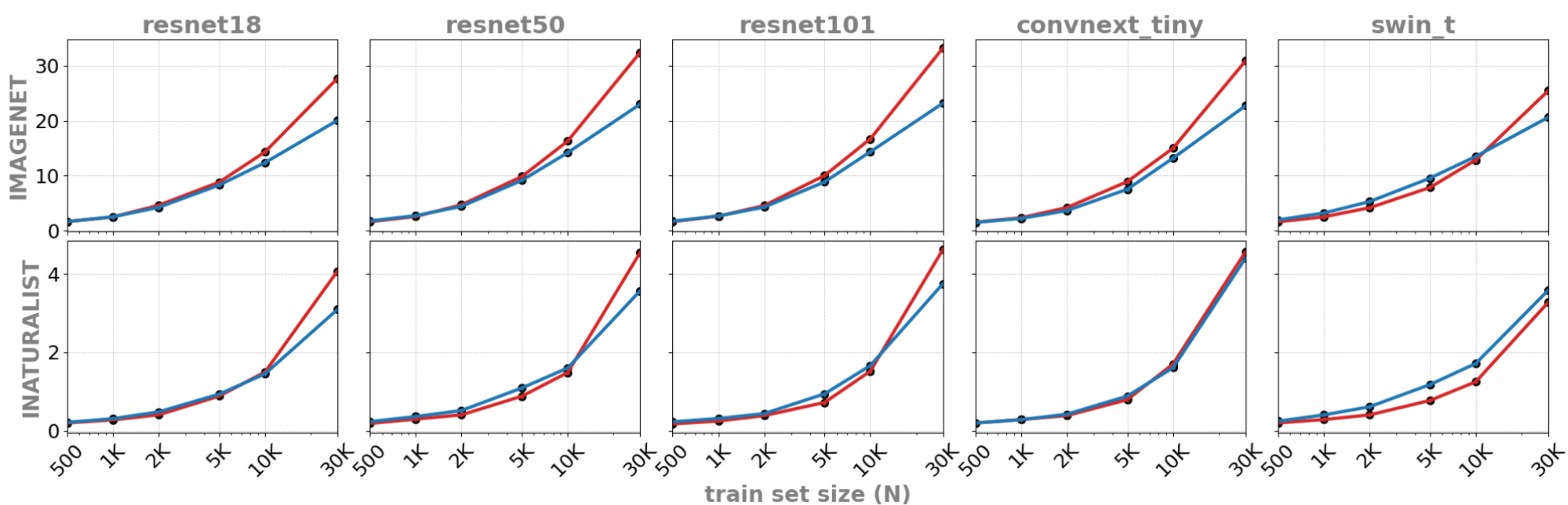}
    \end{minipage}
    \begin{minipage}{\linewidth}
    \caption{\small {\bf DIET matches supervised learning on datasets with only a few samples per class.} Depiction of DIET's downstream performances ({\color{blue} blue}) against supervised learning ({\color{red} red}) controlling training set size ({\bf x-axis}); evaluation is performed over the original full evaluation set. DIET is able to learn highly competitive representations when the dataset is small with only a few samples per classes. See \cref{fig:ablation} for additional datasets.}
    \end{minipage}
    \vspace{-0.3cm}
    \label{fig:ablation_short}
\end{figure*}

{\bf Batch-size does not impact DIET's performance.}~One important question when it comes to training a method with low resources is the ability to employ (very) small batch sizes. This is in fact one reason hindering the deployment of SSL methods which require quite large batch sizes to work (256 is a strict minimum in most cases). Therefore, we perform a small sensitivity analysis in \cref{tab:BS_CV} where we vary the batch size from $8$ to $2048$ without any hyper-parameter tuning other than the standard learning rate scaling used in supervised learning: $lr=0.001\frac{\texttt{bs}}{256}$. We observe small fluctuations of performances (due to a sub-optimal learning rate) but no significant drop in performance, even for batch size of $32$. When going to $16$ and $8$, we observe slightly lower performances, likely due to batch-normalization \cite{ioffe2015batch} which is known to behave erratically below a batch size of $32$ \cite{ioffe2017batch}.
\\
{\bf Data-Augmentation sensitivity is similar to SSL.}~We observed in the previous \cref{sec:SOTA} that when using DA, DIET is able to perform on par with highly engineered state-of-the-art methods. Yet, knowing which DA to employ is not trivial, e.g., many data modalities have no obvious DA. One natural question is, thus, concerning the sensitivity of DIET's performance to the employed DA. To that end, we propose three DA regimes, one only consistent of random crops and horizontal flips ({\bf strength:1}), which could be considered minimal in computer vision, one which adds color jittering and random grayscale ({\bf strength:2}), and one last which further adds Gaussian blur and random erasing \cite{zhong2020random} ({\bf strength:3}); the exact parameters for those transformations are given in \cref{algo:DA}. We observe on TinyImagenet and with a Resnet34 the following performances 32.93$\pm$ 0.6, 45.60$\pm$ 0.2, and  45.75$\pm$ 0.1 respectively over 5 independent runs, details and additional architectures provided in \cref{fig:DA,tab:DA} in the Appendix. We thus observe that while DIET greatly benefit from richer DA (strength:1 $\mapsto$ 2), it however does not require heavier transformation such as random erasing.
\\
{\bf Label smoothing helps.} One important difference in training behavior between supervised learning and SSL is in the number of epochs required to see the quality of the representation plateau. Due to the different loss used in DIET, one might wonder about the differences in training behavior. We observe that DIET takes more epochs than SSL until the loss converges. However, by using large values of label smoothing, {\em e.g.}, $0.8$, it is possible to obtain faster convergence. We provide a sensitivity analysis in \cref{fig:epochs,tab:epochs} in the Appendix. In fact, one should recall that within a single epoch, only one of each datum/class is observed, making the convergence speed of the classifier's {\color{red}$\mW$} matrix the main limitation; we aim to explore improved training strategies in the future as discussed in \cref{sec:conclusion}.


\input{content/conclusion}

\newpage
{
    \small
    \bibliographystyle{unsrt}
    \bibliography{bibliography}
}


\appendix
\onecolumn
\input{content/proof}

\input{content/medmnist}

\newpage

\end{document}

%% file: content/abstract.tex
Deep Learning is often depicted as a trio of data-architecture-loss. Yet, recent Self Supervised Learning (SSL) solutions have introduced numerous additional design choices, e.g., a projector network, positive views, or teacher-student networks. These additions pose two challenges. First, they limit the impact of theoretical studies that often fail to incorporate all those intertwined designs. Second, they slow-down the deployment of SSL methods to new domains as numerous hyper-parameters need to be carefully tuned. In this study, we bring forward the surprising observation that--at least for pretraining datasets of up to a few hundred thousands samples--the additional designs introduced by SSL do not contribute to the quality of the learned representations. That finding not only provides legitimacy to existing theoretical studies, but also simplifies the practitioner's path to SSL deployment in numerous small and medium scale settings. Our finding answers a long-lasting question: the often-experienced sensitivity to training settings and hyper-parameters encountered in SSL come from their design, rather than the absence of supervised guidance.


%% file: content/introduction.tex
\vspace{-0.2cm}
\section{Introduction}
\label{sec:introduction}
\vspace{-0.2cm}

{\em Unsupervised learning} of a model $f_{\vtheta}$, governed by some parameter $\vtheta$, remains a challenging task \cite{bengio2012deep}. In fact, {\em supervised learning} which learns to produce predictions from known input-output pairs can be considered solved in contrast to unsupervised learning which aims to produce useful or intelligible representations from inputs only \cite{hastie2009overview,goodfellow2016deep}. 

{\em Self-Supervised Learning} (SSL) \citep{chen2020simple,misra2020self} has recently demonstrated that one can train, without labels, highly non-trivial Deep Neural Networks (DNNs) whose representations are often richer than supervised ones \citep{zimmermann2021contrastive}. In particular, SSL differs from {\em reconstruction-based} methods such as (denoising, variational, masked) Autoencoders \citep{vincent2008extracting,vincent2010stacked,kingma2013auto} and their variants by removing the need for a {\em decoder} DNN and an input-space reconstruction loss, both being difficult to design \citep{wang2004image,grimes2005bilinear,larsen2016autoencoding,cosentino2022spatial}. 
Nonetheless, SSL which is the current state-of-the-art unsupervised learning solution, comes with many moving pieces, for instance, a carefully designed {\em projector} DNN $g_{\vgamma}$ to perform SSL training with the composition $g_{\vgamma} \circ f_{\vtheta}$ and throwing away the projector ($g_{\vgamma}$) afterwards \cite{chen2020simple}, or advanced anti-collapse techniques involving moving average teacher models \cite{grill2020bootstrap,caron2021emerging}, representation normalization \cite{chen2021exploring,zbontar2021barlow}, or Entropy estimation \cite{chen2020simple,li2021self}. An incorrect pick of any of those moving pieces results in a drastic drop in performances \cite{cosentino2022toward,bordes2022guillotine}. Most of those design choices have, however, been explored, carefully-tuned over many works, and set in stone when considering large scale natural images. {\em But how can one deploy such pipelines to new label-free data modalities when so many design choices need to be carefully tuned?} 

As of today, one would have two solutions. Either avoid learning altogether and use a pretrained model most commonly from Imagenet--which will be highly sub-optimal when considering non natural images such as medical \cite{kim2022transfer}--or cross-validate against the many hyper-parameters of SSL models. Even more limiting, SSL's cross-validation relies on assessing the quality of the produced DNN through the dataset's labels and test accuracy {\em e.g.} from a (supervised) linear probe. This supervised quality assessment is required because current SSL losses fail to convey any qualitative information about the representation being learned \cite{ghosh2022investigating,garrido2022rankme}. Besides the need for labels, SSL's design sensitivity poses a real challenge as current methods are computationally demanding, i.e., they require distributed training on multiple GPUs which, practically, limits the amount of cross-validation that can be performed.

We thus ask the following question. {\em What are the core component of current SSL that are needed to learn strong representations so that (supervised) cross-validation can be thrown-away?} The answer to that question would not only take us closer to a truly unsupervised learning pipeline, but would also help tremendously in the deployment of SSL to new applications, and in our understanding of unsupservised learning.
As we will see, it turns out that---at least for datasets of up to a few hundred thousand samples---many current SSL design choices can be removed altogether without impacting the quality of the final representation. Besides reducing the overall pipeline complexity, our analysis will demonstrate two crucial benefits in stripping down SSL pipelines: (i) the sensitivity of the representation's quality to hyperparameters and architecture changes is greatly improved, i.e., reducing the need for cross-validation, and (ii) the SSL training loss value becomes informative of the quality of the learned representation.
We summarize below the key benefits of employing stripped down SSL pipelines---coined \textbf{DIET}---for reasons that will become clear during our study:
\begin{enumerate}[itemsep=0pt,topsep=0pt]
    \item {\bf Competitive on common benchmarks and SOTA on medical and small datasets:} validated on more than $13$ datasets including natural images (\cref{tab:C100,tab:tiny}) and medical images (\cref{tab:medmnist}) even against SSL benchmarks pretrained on Imagenet (\cref{tab:transfer}).
    \item {\bf Stable and Out-of-the-box:} providing consistently high performances without any hyper-parameter tuning when switching between architectures and datasets, as validated on more than $16$ official architectures including ConvNexts, ViTs, and on 13 datasets (\cref{tab:C100,tab:tiny,tab:transfer}) with the same hyper-parameters (\cref{fig:hparams}), and learning succeeds even with mini-batch sizes as small as 32 (\cref{tab:BS_CV}).
    \item {\bf Data Efficient, single-GPU and theory-friendly:} from the absence of positive pairs, projector networks, and decoders training can be done on single GPU even for high-dimensional images and is suited for theoretical analysis.
    \item {\bf Informative training loss:} the training loss strongly correlates with the downstream task test accuracy across architectures and datasets (\cref{fig:accus}) enabling informed quality assessment of a DIET trained model without requiring labels.
\end{enumerate}

Pseudo-code is provided in \cref{algo:DIET} in addition to an interactive demo  \href{https://colab.research.google.com/drive/1jHzMBYfAVBaIAIAB2d_2CZ-kMCyfhe26?usp=sharing}{colab notebook}. 

%% file: content/background.tex
\vspace{-0.2cm}
\section{Why Self Supervised Learning Needs Occam's Razor}
\label{sec:background}
\vspace{-0.2cm}

Unsupervised learning often takes the form of intricate methods combining numerous moving pieces that need readjustment for each DNN architecture and dataset. As a result reproducibility, transferability across domains, and explainability are hindered.

{\bf Spectral embedding is computationally challenging.}~Spectral embedding takes many forms but can be summarized into estimating geodesic distances \cite{meng2008improving,thomas2013geodesic} between all or some pairs of training samples to then learn a non-parametric \cite{roweis2000nonlinear,belkin2001laplacian,balasubramanian2002isomap,brand2003unifying}, or parametric \cite{bengio2003out,pfau2018spectral} mapping that produces embeddings whose pairwise distances matches the estimated geodesic ones. As such, spectral embedding heavily relies on the estimation of the geodesic distances which is a challenging problem \cite{lantuejoul1981use,lantuejoul1984geodesic,revaud2015epicflow}, especially for images and videos \cite{donoho2005image,wakin2005high}. This limitation motivated the development of alternative methods, \textit{e.g.}, Self-Supervised Learning (SSL) that often employ losses similar to spectral embedding \cite{haochen2021provable,balestriero2022contrastive,cabannes2022minimal} but manage to move away from geodesic distance estimation through the explicit generation of positive pairs, \textit{i.e.}, that are close neighbors on the data manifold.

{\bf Self-Supervised Learning is over-specialized.}~Despite impressive performance and rigorous theoretical motivation, SSL development was mostly driven by industry driven research and thus entirely focused on large-scale natural images and sounds. In fact, SSL has evolved to a point where novel methods are architecture and dataset specific. A few challenges that limit SSL to be widely adopted are (i) loss values which are uninformative of the DNN's quality \cite{reed2021selfaugment,garrido2022rankme}, partly explained by the fact that SSL composes the DNN of interest $f_{\theta}$ with a projector DNN $ g_{\gamma}$ appended to it during training and discarded afterwards, (ii) too many per-loss and per-projector hyper-parameters whose impact on the DNN's performances are hard to control or predict \cite{grill2020bootstrap,tian2021understanding,he2022exploring}, and (iii) lack of transferability of the hyper-parameters across datasets and architectures \cite{zhai2019visual,cosentino2022geometry}Lastly, SSL requires heavy code refactoring, e.g., it requires to generate positive pairs and forward them to siamese DNNs, sometimes with one DNN having parameters as the moving average of the other. 
This makes SSL implementation more costly than supervised learning often requiring distributed training and long training schedules that, effectively, reduce the accessibility and inclusivity of SSL research \citep{crowell2023ai}.

{\bf Reconstruction-based learning is unstable.}~
Reconstruction without careful tuning of the loss has been known to be sub-optimal for long \cite{bishop1994mixture,graves2013generating} and new studies keep reminding us of that \cite{lecun2022path}. The argument is simple, suppose one aims to minimize a reconstruction metric $R$ for some input $\vx$
\begin{align}
    R(d_{\gamma}(e_{\eta}(\vx)),\vx),\label{eq:r}
\end{align}
where $e_{\eta}$ and $d_{\gamma}$ are parametrized learnable encoder and decoder networks respectively; $e_{\eta}(\vx)$ is the representation of interest to be used after training. In practice, as soon as some noise $\epsilon$ is present in the data, \textit{i.e.} we observe $\vx + \epsilon$ and not $\vx$, that noise $\epsilon$ must be encoded by $e_{\eta}$ to minimize the loss from \cref{eq:r} unless one carefully designs $R$ so that 
$R(\vx+\epsilon,\vx) = 0$. However, designing such a {\em noise invariant} $R$ has been attempted for decades \cite{park1995distance,simoncelli1996rotation,fienup1997invariant,grimes2005bilinear,wang2005translation} and remains a challenging open problem. Hence, many solutions rely on learning $R$, e.g., in VAE-GANs \cite{larsen2016autoencoding} bringing even further instabilities and training challenges. Other alternatives carefully tweak $R$ per dataset and architectures, e.g., to only compute the reconstruction loss on parts of the data as with BERT \cite{devlin2018bert} or MAEs \cite{he2022masked}. Lastly, the quality of the encoder representation depends on its architecture but also on the decoder \cite{yang2017improved,xu2021improving} making cross-validation more costly and unstable \cite{antun2020instabilities}.

SSL is the family of method that have produced the most significant state-of-the-art solutions in recent years. Hence, it is the solution of choice that any practitioner hopes to deploy. As such, we propose to take a step towards understanding and alleviating the many practical challenges that would be up against through {\bf DIET}--a stripped down SSL pipeline.

%% file: content/conclusion.tex
\vspace{-0.3cm}
\section{Conclusions and Future Work}
\label{sec:conclusion}
\vspace{-0.2cm}

We examined current SSL pipelines and identified a few core components that clearly improve the quality of learned representations: (i) large number of training epochs, and (ii) strong and informed data augmentation. However, for numerous settings we explored, i.e., dataset with less than a few hundred thousands samples, the additional SSL complications, such as, positive views, nonlinear projector networks, teacher-student networks, do not help. On the contrary, we found that remove those additional parts of SSL pipelines make training much more stable and robust to changes in architecture, data modality, dataset size, and batch size. Even more surprising, the training objective now becomes informative of the downstream tasks test performance. We hope that our findings will help question which parts of our current pipelines are truly needed for case-by-case deployment, when knowing that they have been largely develop for large scale natural image tasks. Another impact of our findings lies the opening new doors to provable learning solutions. In fact, as the simpler pipeline we experimented with is easier to theoretically study, it could help in deriving novel and principled solutions.

%% file: content/proof.tex
\newpage

\begin{center}
    \Huge
    Supplementary Materials
\end{center}

The supplementary materials is providing the proofs of the main's paper formal results. We also provide as much background results and references as possible throughout to ensure that all the derivations are self-contained.
Some of the below derivation do not belong to formal statements but are included to help the curious readers get additional insights into current SSL methods.

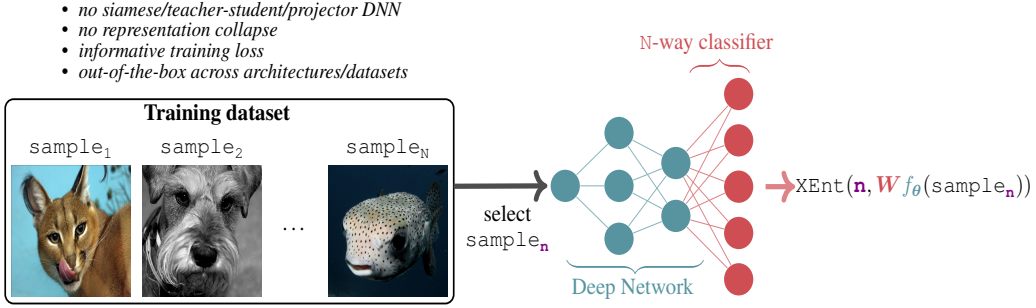
\begin{figure*}[h!]
\centering    
\resizebox{0.99\textwidth}{0.3\textwidth}{%
\begin{tikzpicture}
\node[inner sep=0pt] (a) at (0,0) {\includegraphics[width=2cm,height=2cm]{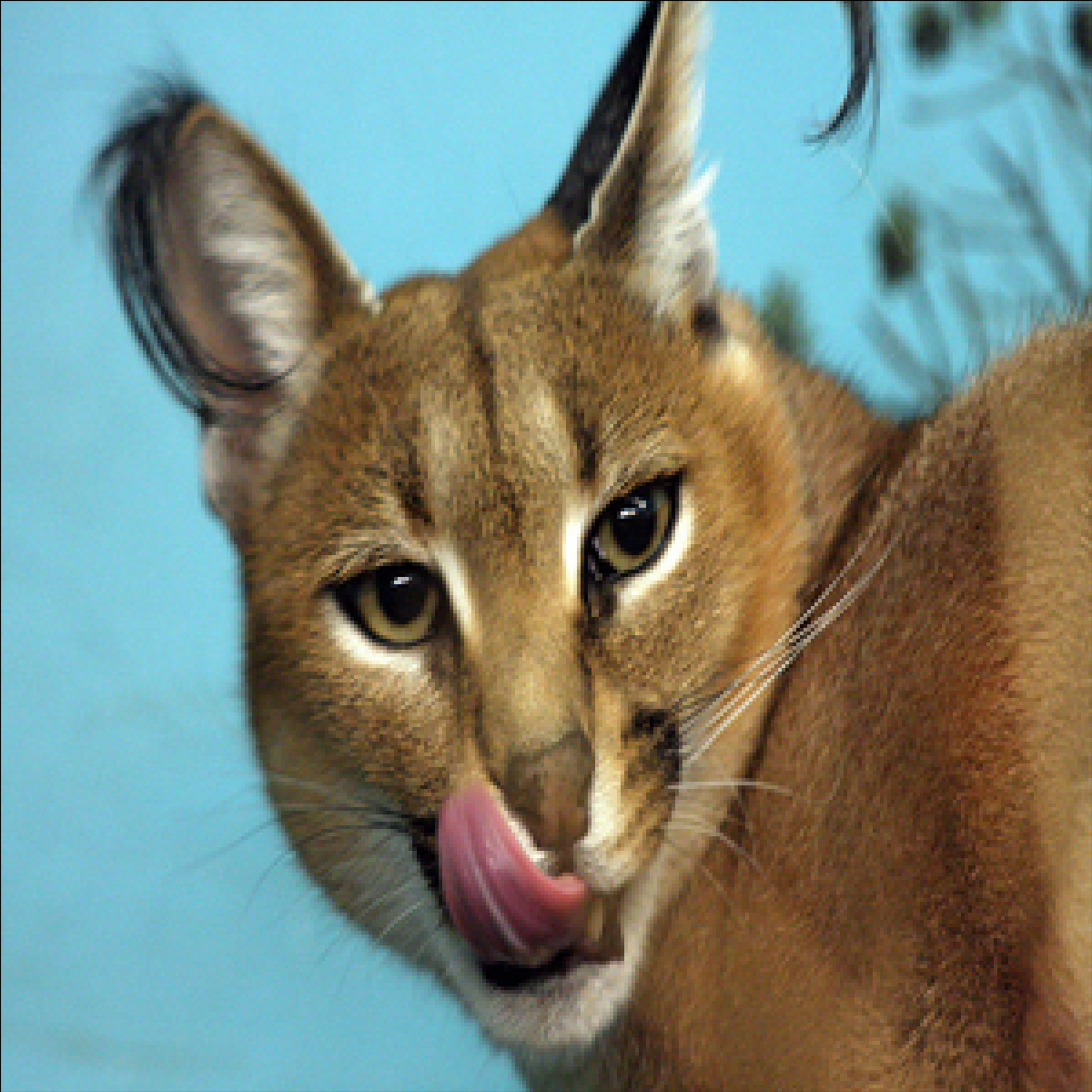}};
\node[inner sep=0pt,above=.1cm of a]  {$\texttt{sample}_{\texttt{1}}$};

\node[inner sep=0pt,right=.2cm of a] (b) {\includegraphics[width=2cm,height=2cm]{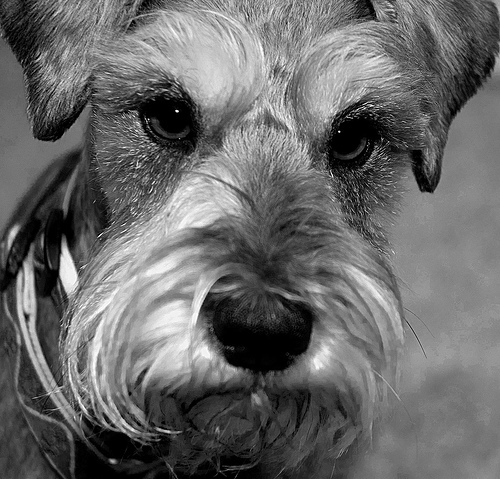}};
\node[inner sep=0pt,above=.1cm of b]  {$\texttt{sample}_{\texttt{2}}$};
\node[inner sep=0pt,above right=.6cm and -2cm of b]  {\textbf{Training dataset}};

\node[inner sep=0pt,right=.2cm of b] (c) {\parbox{0.7cm}{\centering\dots}};

\node[inner sep=0pt,right=.2cm of c] (d) {\includegraphics[width=2cm,height=2cm]{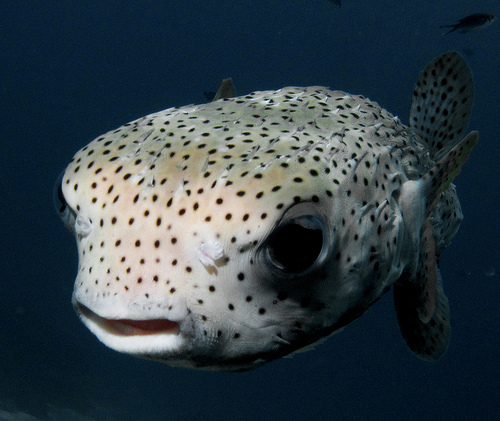}};
\node[inner sep=0pt,above=.1cm of d]  {$\texttt{sample}_{\texttt{N}}$};
    
\draw[black,thick,rounded corners] ($(a.north west)+(-0.1,1)$)  rectangle ($(d.south east)+(0.1,-0.1)$);

\node[inner sep=0pt,below right=-1.4cm and 0.cm of d] (node2) {\parbox{2cm}{\centering select $\texttt{sample}_{\color{violet}\texttt{\textbf{n}}\color{black}}$}};

\definecolor{red}{RGB}{207,78,83}
\definecolor{blue}{RGB}{87,148,160}
\tikzstyle{every pin edge}=[<-,shorten <=1pt]
    \tikzstyle{neuron}=[circle,fill=black!25,minimum size=14pt,inner sep=0pt]
    \tikzstyle{co}=[black!55]
    \tikzstyle{input neuron}=[neuron, fill=blue];
    \tikzstyle{output neuron}=[neuron, fill=red];
    \tikzstyle{hidden neuron}=[neuron, fill=blue];

    \node[input neuron,above right=-0.5cm and 1.8cm of d] (I-2){};

    \node[hidden neuron,right=0.4cm of I-2] (H-3) {};
    \node[hidden neuron,above=0.3cm of H-3] (H-4) {};
    \node[hidden neuron,below=0.3cm of H-3] (H-2) {};
    
    \node[hidden neuron,below right=0.1cm and 0.6cm of H-4] (h-2) {};
    \node[hidden neuron,below=0.3cm of h-2] (h-1) {};

    \node[output neuron, above right=1.5cm and 0.7cm of h-1] (O1) {};

    \foreach \source in {2}
        \foreach \dest in {2,...,4}
            \path[blue!70] (I-\source) edge (H-\dest);
    \foreach \dest in {2,...,4}{
        \path[blue!70] (H-\dest) edge (h-1);
        \path[blue!70] (H-\dest) edge (h-2);
        }
            
    \foreach \source in {1,...,2}
        \path[red!70] (h-\source) edge (O1);
    \foreach \dest/\new in {1/2,2/3,3/4,4/5}{
        \node[output neuron, below=0.2cm of O\dest] (O\new) {};
        \foreach \source in {1,...,2}
            \path[red!70] (h-\source) edge (O\new);
    }
    
\draw [pen colour={red},thick,decorate,decoration={calligraphic brace,amplitude=2mm, raise=1.5mm}] ($(O1.north east)+(-1,0.)$) -- (O1.north east);
\node[red,above right=0.3cm and -2.1cm of O1] {\parbox{2.5cm}{\centering\texttt{N}-way classifier}};

\draw [pen colour={blue},thick,decorate,decoration={calligraphic brace,amplitude=2mm, raise=1.5mm,mirror}] ($(H-2.south west)-(0.7,0)$) -- ($(H-2.south west)+(1.6,0)$);
\node[blue,below right=0.3cm and -1.1cm of H-2] {Deep Network};
    
\node[inner sep=0pt,right=0.7cm of O3] (TO){\texttt{XEnt}$({\color{violet}\texttt{\textbf{n}}},{\color{red}\mW}{\color{blue}f_{\vtheta}}(\texttt{sample}_{\color{violet}\texttt{\textbf{n}}\color{black}})$\color{black}$)$}; 
\path[->,red!70,line width=0.8mm] ($(O3.east)+(0.2,0)$) edge ($(TO.west)+(0,0)$);
\path[->,black!70,line width=0.8mm] ($(d.east)+(0.1,0.68)$) edge ($(I-2)+(-0.3,0)$);


\node[above left=1.1cm and -9cm of a] (top1) {
\parbox{9cm}{\small\em
\begin{itemize}[itemsep=-3pt,topsep=0pt]
    \item no siamese/teacher-student/projector DNN
    \item no representation collapse
    \item informative training loss
    \item out-of-the-box across architectures/datasets
\end{itemize}
}};
\end{tikzpicture}
}
    \caption{\small \textbf{DIET} uses the datum index (\texttt{n}) as the class-target --effectively turning unsupervised learning into a supervised learning problem. In our case, we employ the cross-entropy loss (\texttt{X-Ent}), no extra care needed to handle different dataset or architectures. As opposed to current SOTA, we do not rely on a projector nor positive views \textit{i.e} no change needs to be done to any existing supervised pipeline to obtain DIET. As highlighted in \cref{fig:accus}, DIET's training loss is even informative of downstream test performances, and as ablated in \cref{sec:ablation} there is no degradation of performance with longer training, even for very small datasets (\cref{tab:transfer}).}
    \label{fig:DIET}
\end{figure*}

\section{Linear Model Analysis}
\label{sec:linear_model}

Let's consider the case of a linear model followed by the DIET loss. So the modeling loss given the data matrix $\mX \in \mathbb{R}^{N \times D}$, the linear mapping matrix $\mV \in \mathbb{R}^{D \times K}$ and the DIET linear probe matrix $\mW \in \mathbb{R}^{N \times K}$, is of the form
\begin{align*}
    \mathcal{L} =& \text{CrossEntropy}(\mI, \mX\mV\mW^\top)\\
    =&\sum_{n=1}^{N}-(\mX\mV\mW^\top)_{n,n} + \log\left(\sum_{m=1}^{N}\exp((\mX\mV\mW^\top)_{n,m})\right)\\
    =&\sum_{n=1}^{N}-\langle (\mX\mV)_{n,.},(\mW)_{n,.}\rangle + \log\left(\sum_{m=1}^{N}\exp\left(\langle (\mX\mV)_{n,.},(\mW)_{m,.}\rangle\right)\right),
\end{align*}
the derivative with respect to the parameters $\mV$ and $\mW$ are given by
\begin{align*}
    \nabla_{\mW} = \mA\mX\mV,\;\nabla_{\mV} = \mX^\top\mA\mW
\end{align*}
where the matrix $\mA$ is given by
\begin{align*}
    (\mA)_{i,j} = \left(\frac{e^{(\mX\mV\mW^\top)_{i,j}}}{\sum_{n=1}^{N}e^{(\mX\mV\mW^\top)_{i,n}}}-1_{\{i=j\}}\right).
\end{align*}
The above analysis is true for any matrix $\mX, \mV, \mW$. Finding a general solution by setting the gradient to $0$ is not trivial due to the $\mA$ matrix involving a softmax operation. However, for a special class of data matrices $\mX$, we are able to find a closed-form optimal solution for $\mV$ and $\mW$.
Let's now consider the following low-rank model for the input data matrix $\mX$ as
\begin{align*}
    \mX \triangleq \begin{bmatrix}
        \mu_1,
        \dots ,
        \mu_1,
        \dots,
        \mu_{K},
        \dots ,
        \mu_{K}
    \end{bmatrix}^\top,
\end{align*}
where each $\mu_i$ is repeated $N/K$ times. That is, we assume that $\mX$ has a low-rank structured made of ``centroids''. Note that while we assume here that each centroid is repeated the same number of times to simplify notations, none of the following results require uniform distribution of the centroids. 

\begin{figure}[t!]
    \centering
    \includegraphics[width=0.8\linewidth,height=5cm]{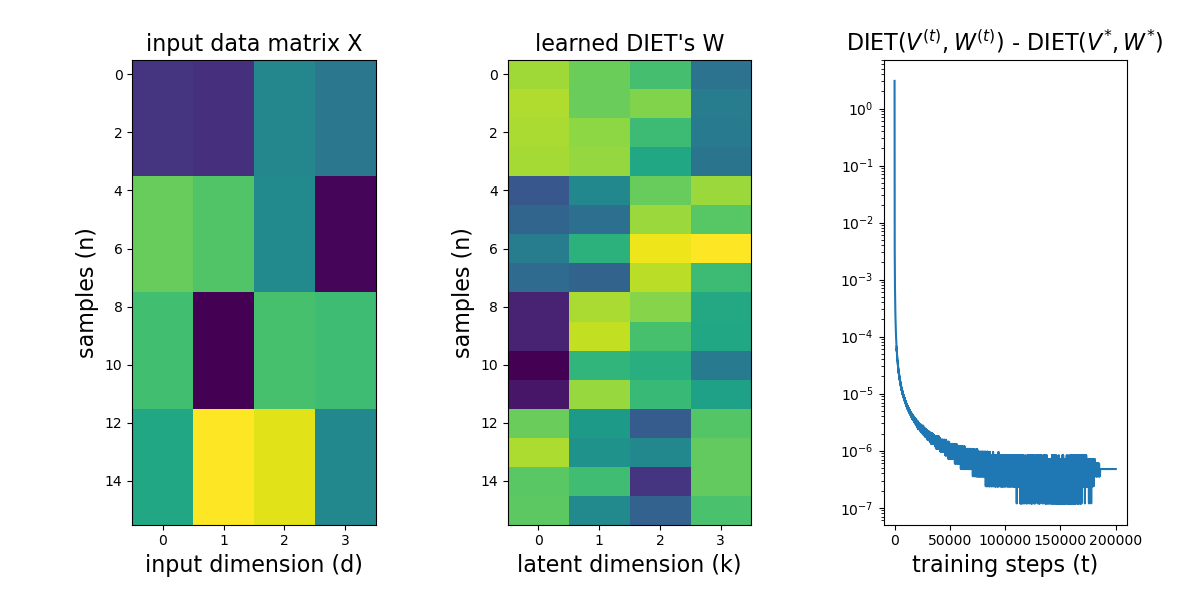}
    \caption{Empirical validation of \cref{sec:linear_model} depicting the optimal solution for DIET for the parameters $\mW$ and $\mV$ under a clustered input data assumption ({\bf left column}), in this case, made of four clusters with four samples per cluster. The learned $\mW$ given in the {\bf middle column} converge to the same clustering, as predicted by our closed-form solution. We also obtain in the {\bf right column} the evolution of the DIET training loss that we compare against the optimal value of the loss (obtained from the optimal parameters). We see that the training converges towards the optimal value of the loss (up to 1e-7 at the end of that training episode).}
    \label{fig:diet_optimal}
\end{figure}

\begin{figure}[t!]
    \centering
    \begin{minipage}{0.5\linewidth}
        
    \includegraphics[width=\linewidth]{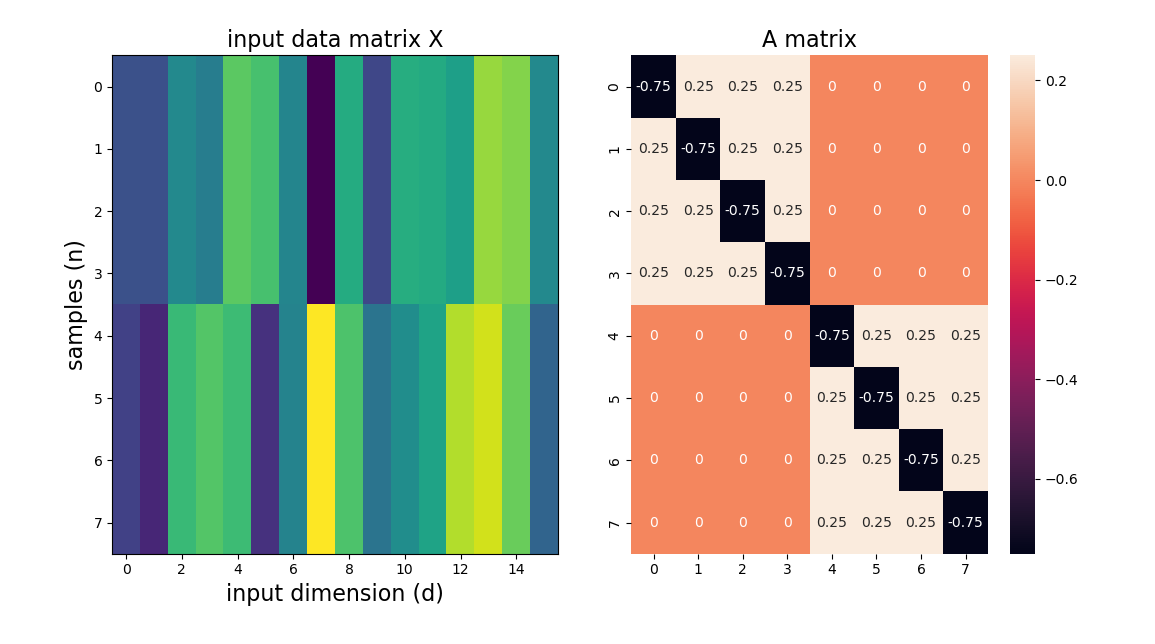}
    \end{minipage}
    \begin{minipage}{0.29\linewidth}
    \caption{Depiction of the optimal $\mA$ matrix (recall \cref{eq:A}) on the {\bf right}, obtained empirically from inserting the optimal parameters that we found for $\mW$ and $\mV$. As predicted by \cref{eq:A} that matrix is made of blocks aligned with the original clustering of the input data matrix $\mX$ given on the {\bf left}.}
    \label{fig:diet_A}
    \end{minipage}
\end{figure}

Then, DIET will effectively learn the clustering, as per the SVD of the data. In fact, let's consider the following parameters $\mV = \mV_{\mX}\mSigma_{\mX}^{-1}$ and $\mW=\kappa \mU_{\mX}$ where we used the (reduced) singular value decomposition of $\mX$ as $\mX = \mU_{\mX}\mSigma_{\mX}\mV_{\mX}^\top$, and with $\kappa \gg 0$. In that setting, the matrix $\mA$ becomes with a block structure as per
\begin{align}
    (\mA)_{i,j} = \frac{1}{K}1_{\{[i/(N/K)]=[j/(N/K)]\}}-1_{\{i=j\}},\label{eq:A}
\end{align}
as depicted in \cref{fig:diet_A}. This leads to a zero-gradient 
\begin{align*}
    \nabla_{\mW} = \mathbf{0},\nabla_{\mV} = \mathbf{0},
\end{align*}
effectively showing that we obtain the optimal parameters, as depicted in \cref{fig:diet_optimal}.

\newpage

\section{Code}
\label{sec:code}

\begin{figure}[t!]
\begin{framed}
\small
    \textbf{DIET's experimental setup:}
    \begin{itemize}[leftmargin=3.5mm,itemsep=-1pt,topsep=0.pt]
        \item \color{blue}Official Torchvision architectures \color{black} (\ul{no changes in init./arch.}), only swapping the classification layer with \color{red}DIET's one \color{black} (right of \cref{fig:DIET}), \ul{no projector DNN}
        \item \ul{Same DA pipeline} ($\mathcal{T}$ in \cref{fig:DIET}) across datasets/architectures with \ul{batch size of 256} to fit on 1 GPU
        \item \ul{AdamW optimizer with linear warmup (10 epochs) and cosine annealing} learning rate schedule, \ul{\texttt{XEnt} loss} (right of \cref{fig:DIET}) with \emph{label smoothing of $0.8$} 
        \item \emph{Learning rate/weight-decay} of $0.001/0.05$ for non transformer architectures and $0.0002/0.01$ for transformers
    \end{itemize}
    \vspace{-0.2cm}
    \caption{In \ul{underlined} are the design choices directly ported from standard supervised learning (not cross-validated for DIET), in \emph{italic} are the design choices cross-validated for DIET but held constant across this study unless specified otherwise. Batch-size sensitivity analysis is reported in \cref{tab:BS_CV,fig:BS_CV} showing that performances do not vary when taking values from $32$ to $4096$. \texttt{XEnt}'s label smoothing parameter plays a role into DIET's convergence speed, and is cross-validated in \cref{fig:epochs,tab:epochs}; we also report DA ablation in \cref{fig:DA,tab:DA}.}
    \label{fig:hparams}
    \end{framed}
    \vspace{-0.5cm}
\end{figure}









\begin{algorithm}[h]
\begin{lstlisting}[language=Python,escapechar=\%,numbers=none]
model = torchvision.models.__dict__[architecture]()

# CIFAR procedure to adjust to the lower image resolution
if is_cifar and "resnet" in architecture:
    model.conv1 = torch.nn.Conv2d(3, 64, kernel_size=3, stride=1, padding=2, bias=False)
    model.maxpool = torch.nn.Identity()

# for each architecture, remove the classifier and get the output dim. (K)
if "alexnet" in architecture:
    K = model.classifier[6].in_features
    model.classifier[6] = torch.nn.Identity()
elif "convnext" in architecture:
    K = model.classifier[2].in_features
    model.classifier[2] = torch.nn.Identity()
elif "convnext" in architecture:
    K = model.classifier[2].in_features
    model.classifier[2] = torch.nn.Identity()
elif "resnet" in architecture or "resnext" in architecture or "regnet" in architecture:
    K = model.fc.in_features
    model.fc = torch.nn.Identity()
elif "densenet" in architecture:
    K = model.classifier.in_features
    model.classifier = torch.nn.Identity()
elif "mobile" in architecture:
    K = model.classifier[-1].in_features
    model.classifier[-1] = torch.nn.Identity()
elif "vit" in architecture:
    K = model.heads.head.in_features
    model.heads.head = torch.nn.Identity()
elif "swin" in architecture:
    K = model.head.in_features
    model.head = torch.nn.Identity()
\end{lstlisting}
\caption{\small Get the output dimension and remove the linear classifier from a given torchvision model (Pytorch used for illustration).}
\label{algo:architectures}
\end{algorithm}

\subsection{Pushing the DIET to Large Models and Datasets}
\label{sec:scaling}

Given DIET's formulation of considering each datum as its own class, it is natural to ask ourselves how scalable is such a method. Although we saw that on small and medium scale dataset, DIET's was able to come on-par with most current SSL methods, it is not clear if this remains true for larger datasets. In this section we briefly describe what can be done to employ DIET on datasets such as Imagenet and INaturalist.

The first dataset we consider is INaturalist which contains slightly more than $500K$ training samples for its mini version (the one commonly employed, see {\em e.g. }\cite{zbontar2021barlow}). It contains almost $10K$ actual classes and most SSL methods focus on transfer learning {\em e.g.} transferring with a Resnet50 from Imagenet-1k lead to SimCLR's 37.2$\%$, MoCoV2's 38.6, BYOL's 47.6 and BarlowTwins'46.5. However training on INaturalist directly produces lower performances reaching only 29.1 with MSN and a ViT. Using DIET is possible out-of-the-box with Resnet18 and ViT variants as their embedding is of dimension 512 and 762 respectively making $\mW$ fit in memory. We obtain 22.81 with a convnext small, and 21.6 with a ViT.

The second dataset we consider is the full Imagenet-1k dataset which contains more than 1 million training samples and 1000 actual classes. In this case, it is not possible to directly hold $\mW$ in-memory. We however tried a simple strategy which simply consists of sub-sampling the training set to a more reasonable size. This means that although we are putting aside many training images, we enable single GPU Imagenet training with DIET. With a training size of $400K$, we able to reach 44.05 with a convnext small, 43.78 with a SwinTiny, and 44.89 with a ViT/B/16. A standard SSL pipeline has performances ranging between $64\%$ and $72\%$. From those experiments, it is clear that DIET's main limitation comes from very large training set sizes. Although the above simple strategy offers a workable solution, it is clearly not sufficient to match with existing unsupervised learning method and thus should require further consideration. As highlighted in \cref{sec:conclusion} below, this is one key avenue for future work.

\begin{figure}[t!]
    \centering
    \includegraphics[width=\linewidth]{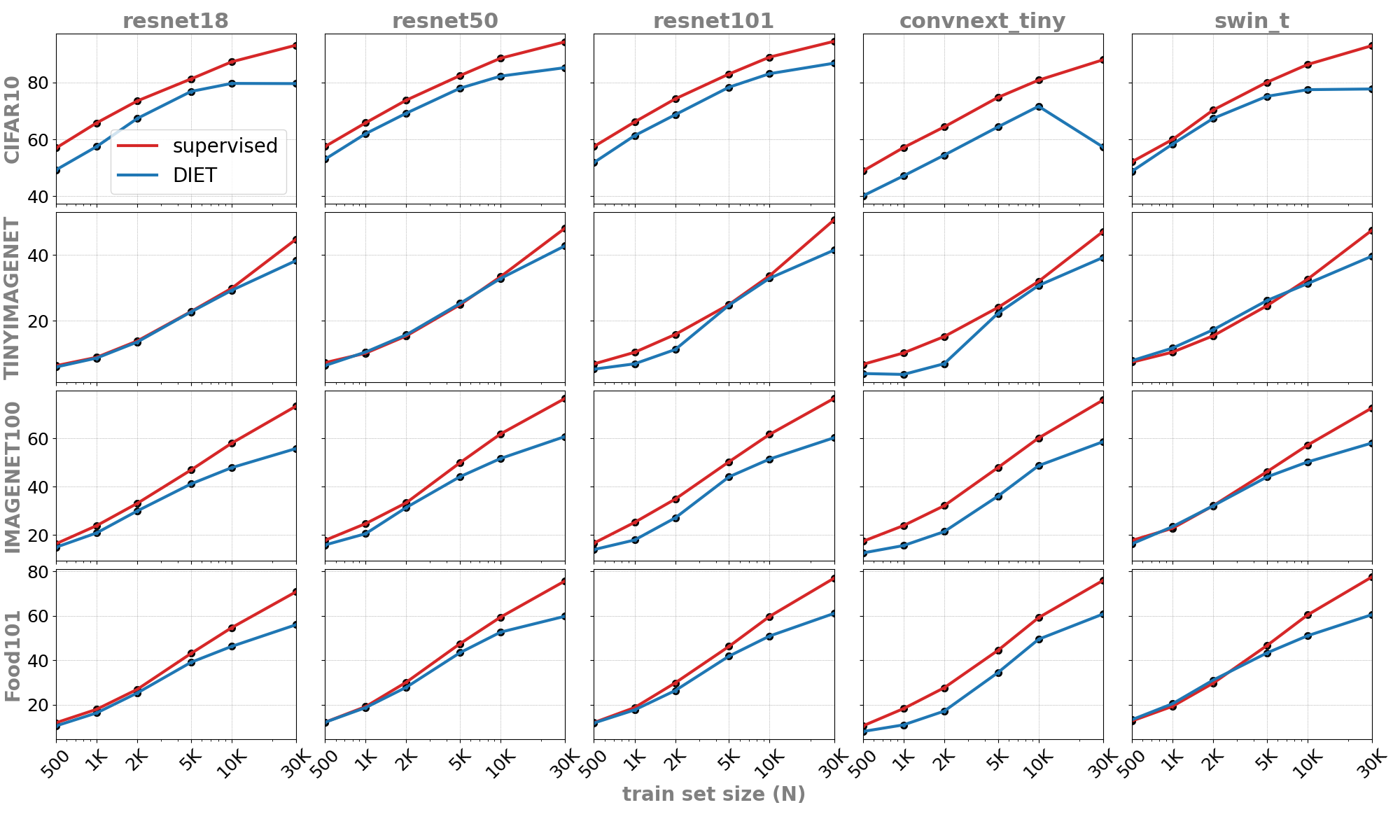}
    \caption{Reprise of \cref{fig:ablation_short} on additional datasets depicting how DIET is able to compete with supervised learning for in-distribution generalization in very small dataset regime.}
    \label{fig:ablation}
\end{figure}

\section{Impact of Training Time and Label Smoothing}

\begin{figure}[h]
    \centering
    Resnet18\\
    \includegraphics[width=\linewidth,height=3cm]{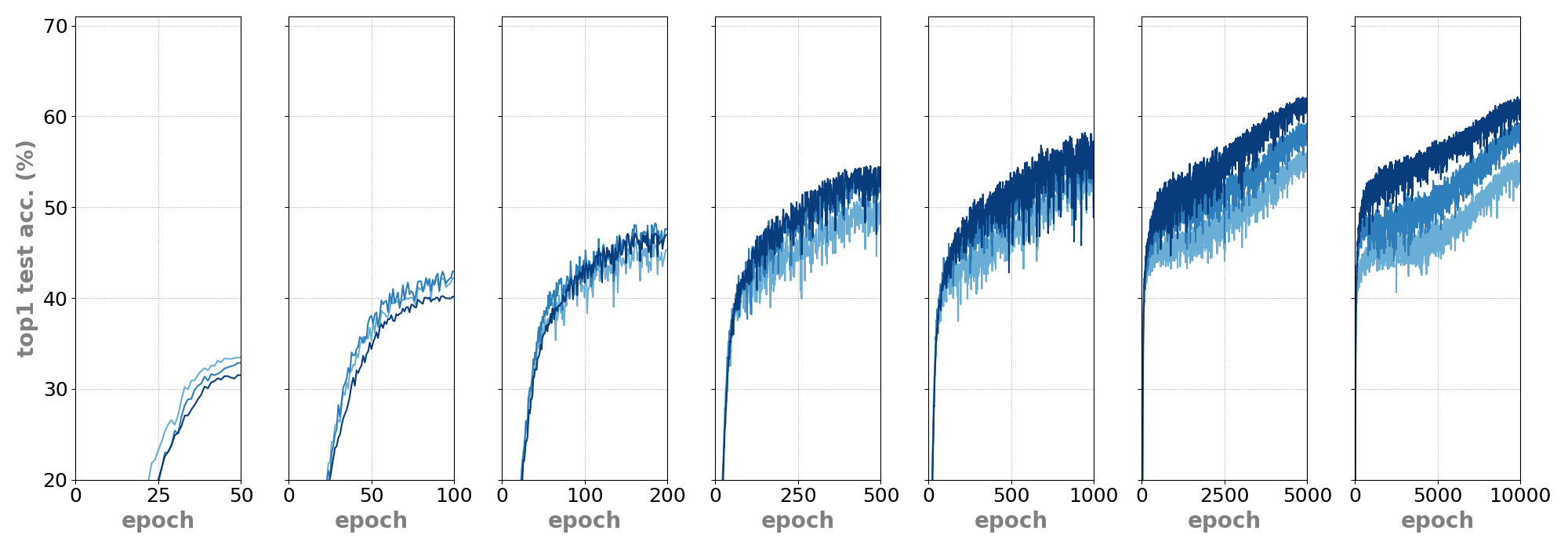}\\
    Resnet50\\
    \includegraphics[width=\linewidth,height=3cm]{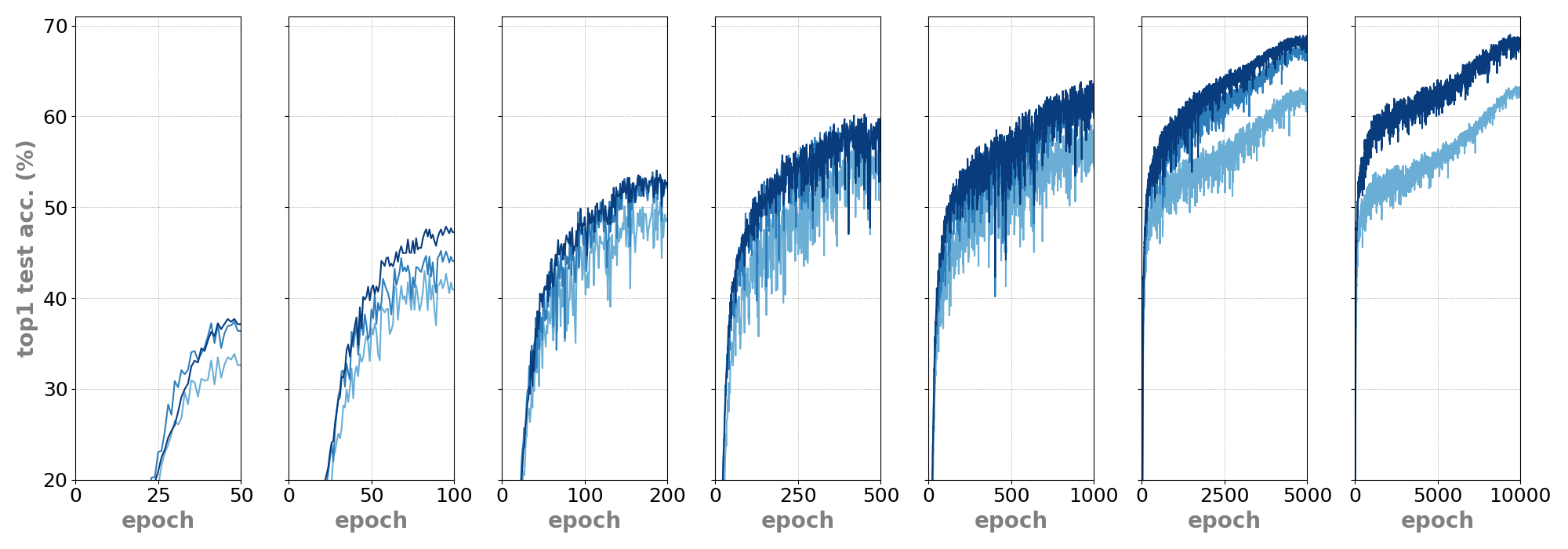}\\
    Resnet101\\
    \includegraphics[width=\linewidth,height=3cm]{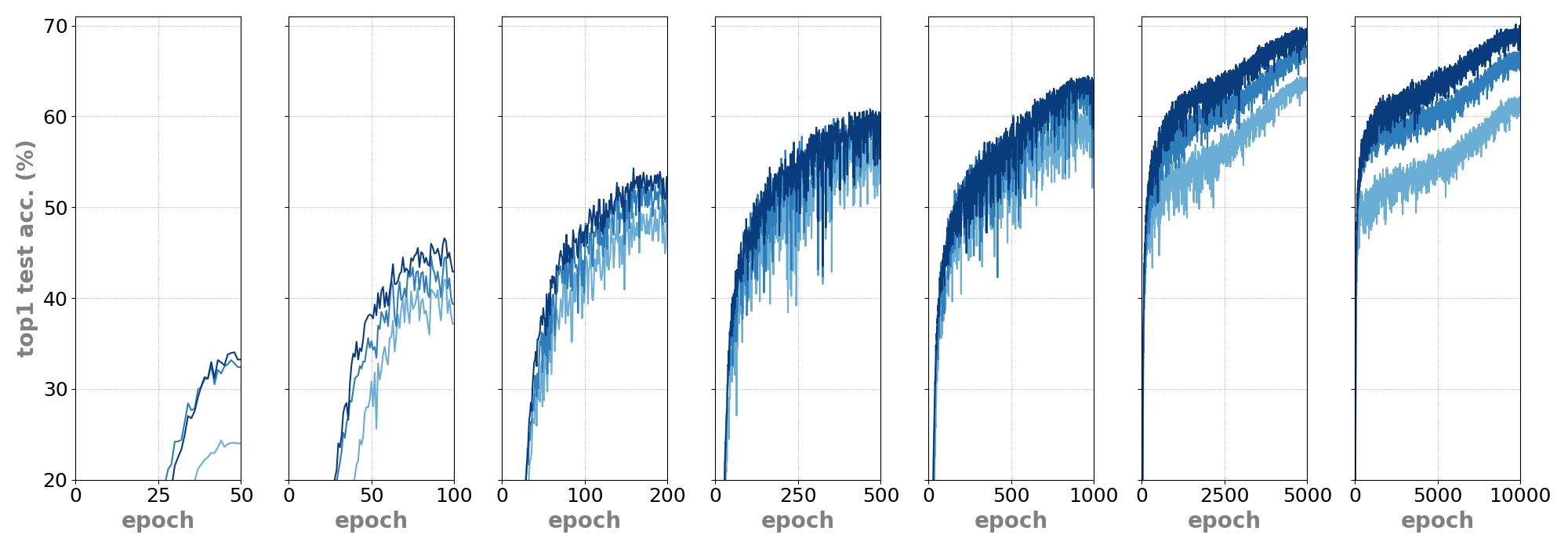}
    \caption{Depiction of the evolution of linear top1 accuracy throughout epochs on CIFAR100 with three Resnet variants and three label smoothing parameters represented by the different {\bf shades of blue} going from light to dark shades with values of $0.1$, $0.4$, and $0.8$ respectively. }
    \label{fig:epochs}
\end{figure}

In Figure \ref{fig:epochs} we show the performance of DIET on CIFAR100 across three label smoothing settings. We find higher values of label smoothing speed up convergence, although in this setting all cases greatly benefit from longer training schedules; final linear probe performances are reported in \cref{tab:epochs}.

\section{Impact of Mini-Batch Size}

\begin{figure}[h]
    \centering
    \begin{minipage}{0.49\linewidth}
    \includegraphics[width=\linewidth]{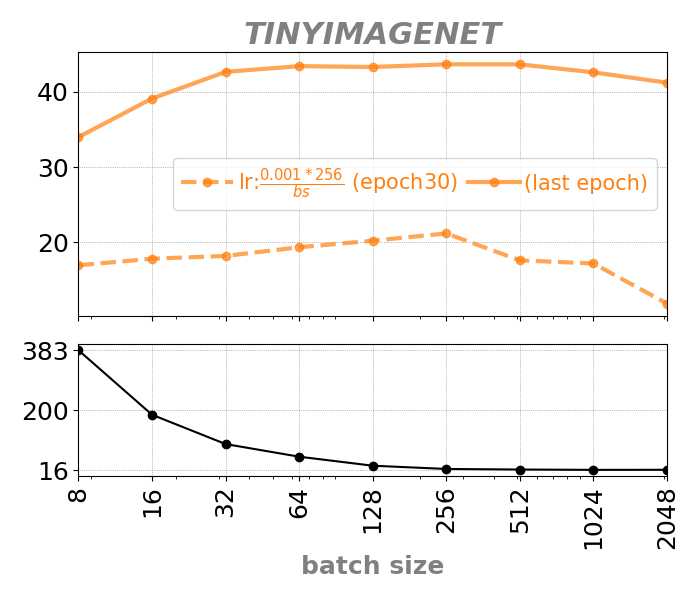}
    \end{minipage}
    \begin{minipage}{0.49\linewidth}
    \includegraphics[width=\linewidth]{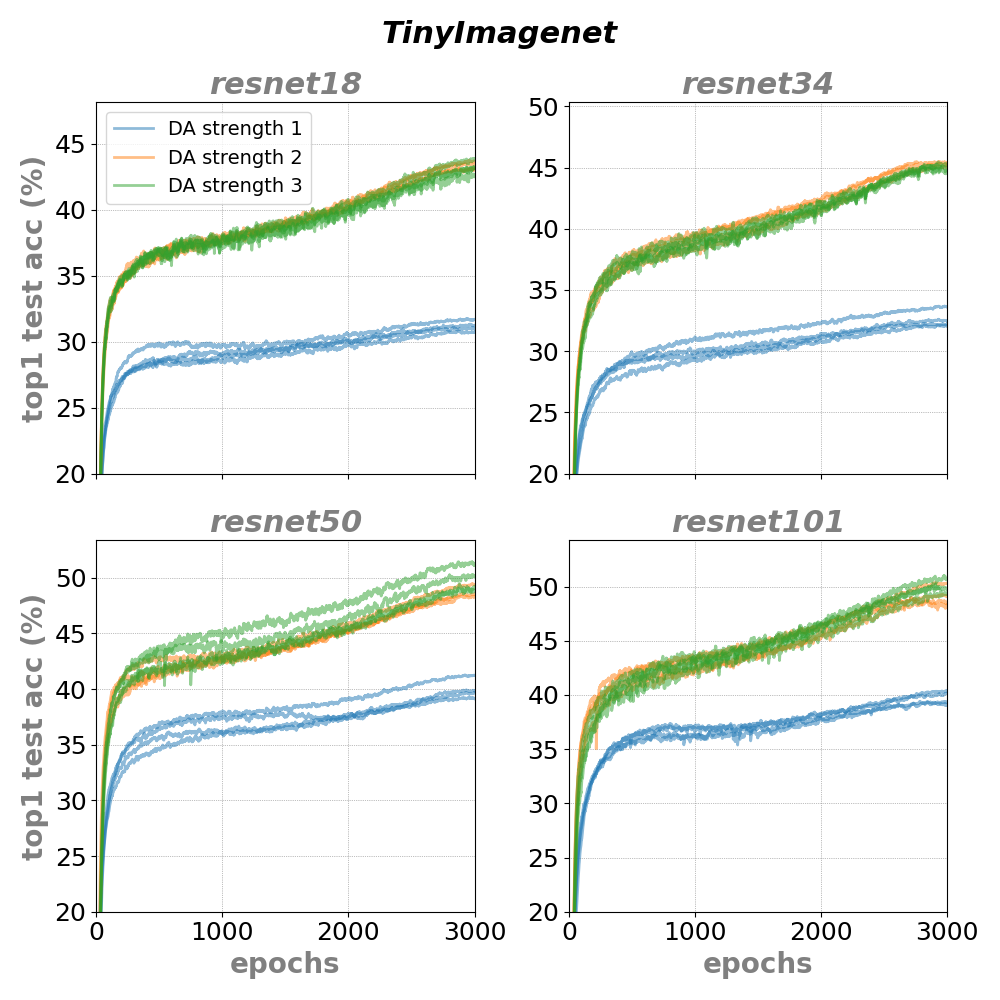}
    \end{minipage}
    \caption{{\bf Left:}TinyImagenet with fixed number of epochs and a single learning rate which is adjusted for each case using the LARS rule therefore per batch-size learning cross-validation can only improve performances, see \cref{tab:BS_CV}, , the per-epoch time includes training, testing, and checkpointing. {\bf Right:} TinyImagenet, see \cref{tab:DA} for table of results, and the specific DAs can be found in \cref{algo:DA}.}
    \label{fig:BS_CV}
    \label{fig:DA}
\end{figure}

We show in \ref{fig:BS_CV} ablations for TinyImagenet using DIET. In addition we show DIET"s robustness to batch size by conducting an additional ablation by varying the batch size for the Derma MedMNIST dataset with batch sizes as low as 8. As shown in Table \ref{app:BS_CV_medmnist}, we see DIET performs well even with very small batch sizes.

\begin{table}[t!]
\centering
\resizebox{0.5\columnwidth}{!}{%
\begin{tabular}{lccccc}
\rowcolor[HTML]{FFFFFF} 
\hline 
\textbf{Batch   Size} & \textbf{8} & \textbf{32} & \textbf{64} & \textbf{128} & \textbf{512} \\
\hline 
DIET                  & 71.87      & 72.52       & 73.07       & 74.36        & 71.02        \\
MoCov2                & 66.88      & 64.64       & 66.73       & 66.88        & 61.40        \\
SimCLR                & 63.14      & 66.43       & 66.83       & 66.88        & 66.83        \\
VICReg                & 65.84      & 60.45       & 64.79       & 66.78        & 66.88       \\
\hline 
\end{tabular}%
}
\caption{Reprise of \cref{tab:BS_CV}: DIET's performance across varying batch sizes on the Derma MedMNIST dataset with all other hyperparameter fixed demonstrating the stability of DIET do that hyper-parameter and across training iterations. All models are trained for 500 epochs.}
\label{app:BS_CV_medmnist}
\end{table}

\section{Impact of Data-Augmentation}

To further study the effect of data augmentation in DIET we study varying data augmentation strengths for TinyImageNet in \cref{fig:DA}. We also examine the effect of weaker data augmetnations for smaller medical images using PathMNIST in \cref{app-tab:medmnist-diet-vit-compare-augmentations}.

\begin{algorithm}[h]
\begin{lstlisting}[language=Python,escapechar=\%,numbers=none]
transforms = [
    RandomResizedCropRGBImageDecoder((size, size)),
    RandomHorizontalFlip(),
]
if strength > 1:
    transforms.append(
        T.RandomApply(
            torch.nn.ModuleList([T.ColorJitter(0.4, 0.4, 0.4, 0.2)]), p=0.3
        )
    )
    transforms.append(T.RandomGrayscale(0.2))
if strength > 2:
    transforms.append(
        T.RandomApply(
            torch.nn.ModuleList([T.GaussianBlur((3, 3), (1.0, 2.0))]), p=0.2
        )
    )
    transforms.append(T.RandomErasing(0.25))
\end{lstlisting}
\caption{\small Custom dataset to obtain the indices ($n$) in addition to inputs $\vx_n$ and (optionally) the labels $y_n$ to obtain \texttt{train\_loader} used in \cref{algo:DIET} (Pytorch used for illustration).}
\label{algo:DA}
\end{algorithm}

\section{DIET compared to supervised learning}
\label{app:diet_v_supervised}

{\bf DIET matches supervised learning on datasets with only a few samples per class.}~
In \cref{fig:ablation_short} we directly compare DIET with supervised learning on a variety of models and datasets but with controlled training size. We clearly observe that for small dataset, {\em i.e.}, for which we only use a small part of the original training set (less than $30$ images per class), DIET's learned representation is as efficient as the supervised one for the in-distribution classification downstream task.

\paragraph{DIET works with scattering network architectures}
As an additional test, scattering networks \citep{oyallon2018scattering,gauthier2022parametric} hard-code part of the model parameters to be wavelet filter-banks. That specification naturally makes such scattering networks very competitive for small data regimes since the number of degrees of freedom is reduced. We therefore performed two additional experiments: Training a hybrid scattering network in a supervised setting Training a hybrid scattering network with DIET and then learning a linear probe on top (keeping the hybrid scattering frozen) We perform both cases above on the full CIFAR10 training set and on a reduced training set of 5000 ($10\%$ of the training data) samples.
Supervised training of the scattering network results in $72.1\%$ ($58.2\%$) test set accuracy, whereas unsupervised DIET pretraining followed by a linear probe results in $77.64\%$ ($62.8\%$) for the same architecture.
From that experiment we obtain two novel insights. First, DIET works out-of-the-box on DNs such as the hybrid scattering network, with a reduced number of parameters. Second, even in that regime, DIET provides strong performances.

%% file: content/medmnist.tex
\section{Additional Results for MedMNIST}
\label{app:medmnist}

\begin{figure}[t!]
    \centering
    \includegraphics[width=0.49\textwidth]{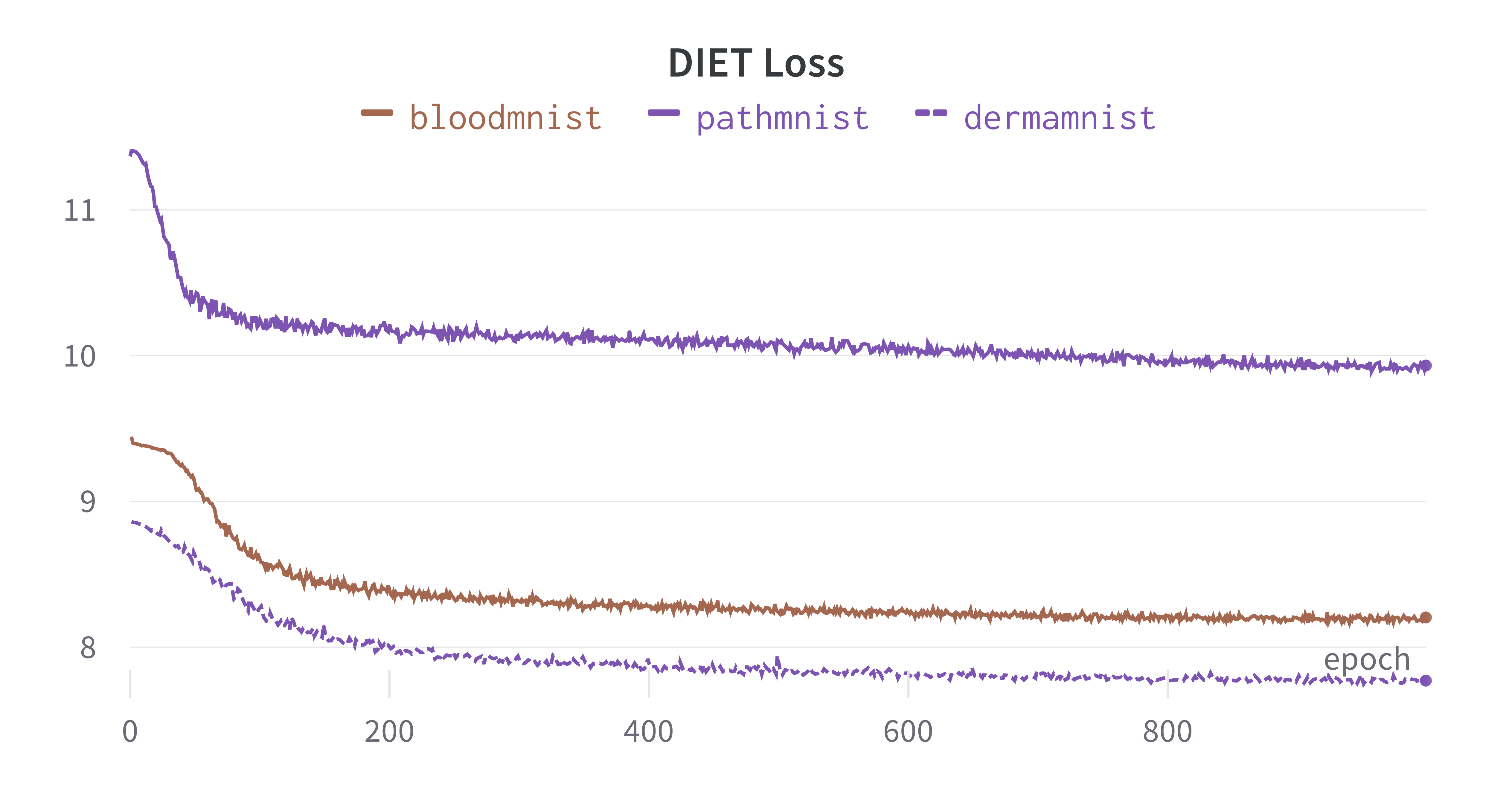}
    \includegraphics[width=0.49\textwidth]{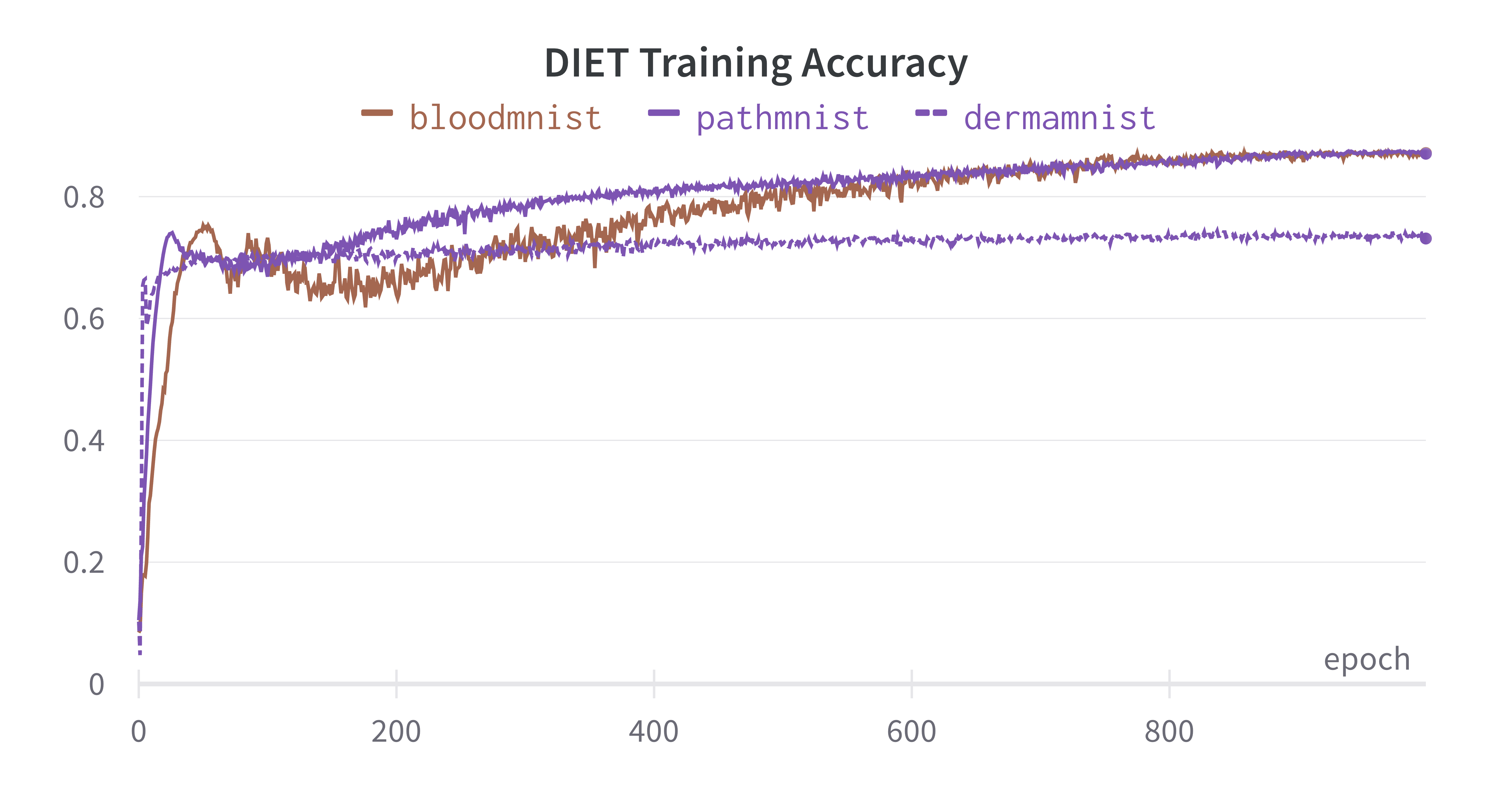}
    \caption{DIET MedMNIST training loss curves for the DIET criterion (left) and training accuracy (right) with a ResNet18 backbone.}
    \label{fig:medmnist-diet-loss}
\end{figure}

In Figure \ref{fig:medmnist-diet-loss} we show training curves for DIET with a ResNet18 architecture. We perform additional experiments with DIET using a vision transformer architecture (ViT-Small with patch size 4) based on the architecture from \url{https://github.com/lucidrains/vit-pytorch/blob/main/vit_pytorch/vit_for_small_dataset.py}. We find DIET achieves good performance on the same MedMNIST datasets with this ViT architecture without additional hyperparameter tuning as shown in Table \ref{app-tab:medmnist-diet-vit} and in comparison to all three baseline SSL methods in Table \ref{app-tab:medmnist-diet-vit-compare-baselines}. 

We find evidence of the default augmentations for PathMNIST being too aggressive and confirm DIET's performance improves with the use of weaker augmentations in Table \ref{app-tab:medmnist-diet-vit-compare-augmentations}. Surprisingly, we find DIET performs quite well with no augmentations at all, a setting in which most standard SSL methods would be impossible to train.


\begin{table}[t!]
\centering
\begin{minipage}{0.6\linewidth}
\centering
    \begin{tabular}{lcccccc}
\hline
\textbf{} & \multicolumn{2}{c|}{\textbf{bloodmnist}}            & \multicolumn{2}{c|}{\textbf{dermamnist}}            & \multicolumn{2}{c}{\textbf{pathmnist}} \\
\textbf{} & \textbf{train} & \multicolumn{1}{c|}{\textbf{test}} & \textbf{train} & \multicolumn{1}{c|}{\textbf{test}} & \textbf{train}     & \textbf{test}     \\ \hline
\textbf{DIET}   & 77.65 & 81.85 & 71.03 & 68.88 & 56.37  & 21.27 \\
\textbf{SimCLR} & 82.48 & 79.45 & 69.13 & 32.37 & 69.45 & 21.80  \\
\textbf{VICReg} & 86.71 & 81.03 & 69.89 & 46.33 & 82.94 & 12.76 \\
\textbf{MoCov2} & 62.76 & 51.01 & 66.78 & 63.39 & 72.9  & 41.75 \\
\hline\\
\end{tabular}%
\end{minipage}
\begin{minipage}{0.6\linewidth}
\centering
\begin{tabular}{lcc}
\hline
DIET         & \multicolumn{2}{c}{PathMNIST} \\
\hline
Augmentation & train         & test          \\
Default      & 56.37         & 21.27         \\
Weak         & 44.90         & 48.95         \\
None         & 44.65         & 45.67        \\
\hline
\end{tabular}
\end{minipage}
    \caption{{\bf Top:}DIET performance across the three MedMNIST datasets using a transformer (ViT-S) architecture with patch size 4 in comparison to standard SSL baselines with the same ViT architecture. {\bf Bottom:}Comparing DIET's performance across data augmentations for PathMNIST using a transformer (ViT-S) architecture with patch size 4. Weak augmentation corresponds to only random resized cropping and horizontal flipping.}
    \label{app-tab:medmnist-diet-vit-compare-baselines}
    \label{app-tab:medmnist-diet-vit-compare-augmentations}
\end{table}

\begin{table}[t!]
    \setlength{\tabcolsep}{0.4em}
\centering
    \caption{\small DIET performance across the three MedMNIST datasets using a transformer (ViT-S) architecture with patch size 4. In the first row we show the performance of a baseline
    SimCLR model with the default ResNet18 encoder for comparison.
    }
    \label{app-tab:medmnist-diet-vit}
\begin{tabular}{lcccccc}
\hline
dataset & \multicolumn{2}{c|}{bloodmnist}   & \multicolumn{2}{c|}{dermamnist}   & \multicolumn{2}{c}{pathmnist} \\
        & train & \multicolumn{1}{c|}{test} & train & \multicolumn{1}{c|}{test} & train         & test          \\ \hline
DIET    & {\color{gray}77.65} & 81.85                     & {\color{gray}71.03} & 68.88                     & {\color{gray}56.37}         & 21.27         \\ \hline
\end{tabular}%
\end{table}



\newpage
\section*{NeurIPS Paper Checklist}

\begin{enumerate}

\item {\bf Claims}
    \item[] Question: Do the main claims made in the abstract and introduction accurately reflect the paper's contributions and scope?
    \item[] Answer: \answerYes{}
    \item[] Justification: \answerNA{}

\item {\bf Limitations}
    \item[] Question: Does the paper discuss the limitations of the work performed by the authors?
    \item[] Answer: \answerYes{} 
    \item[] Justification: \answerNA{}

\item {\bf Theory Assumptions and Proofs}
    \item[] Question: For each theoretical result, does the paper provide the full set of assumptions and a complete (and correct) proof?
    \item[] Answer: \answerYes{} 
    \item[] Justification: \answerNA{}

    \item {\bf Experimental Result Reproducibility}
    \item[] Question: Does the paper fully disclose all the information needed to reproduce the main experimental results of the paper to the extent that it affects the main claims and/or conclusions of the paper (regardless of whether the code and data are provided or not)?
    \item[] Answer: \answerYes{} 
    \item[] Justification: \answerNA{}

\item {\bf Open access to data and code}
    \item[] Question: Does the paper provide open access to the data and code, with sufficient instructions to faithfully reproduce the main experimental results, as described in supplemental material?
    \item[] Answer: \answerYes{} 
    \item[] Justification: \answerNA{}

\item {\bf Experimental Setting/Details}
    \item[] Question: Does the paper specify all the training and test details (e.g., data splits, hyperparameters, how they were chosen, type of optimizer, etc.) necessary to understand the results?
    \item[] Answer: \answerYes{} 
    \item[] Justification: \answerNA{}

\item {\bf Experiment Statistical Significance}
    \item[] Question: Does the paper report error bars suitably and correctly defined or other appropriate information about the statistical significance of the experiments?
    \item[] Answer: \answerNo{} 
    \item[] Justification: We do not report eror bars, but instead carefully study and report the stability of our results across various hyperparameter and architecture choices to make clear the results are not an artifact of stochasticity during training. For baselines, we report numbers from publicly available papers when possible, which we found often lack error bars as well.

\item {\bf Experiments Compute Resources}
    \item[] Question: For each experiment, does the paper provide sufficient information on the computer resources (type of compute workers, memory, time of execution) needed to reproduce the experiments?
    \item[] Answer: \answerYes{}
    \item[] Justification: \answerNA{}
    
\item {\bf Code Of Ethics}
    \item[] Question: Does the research conducted in the paper conform, in every respect, with the NeurIPS Code of Ethics \url{https://neurips.cc/public/EthicsGuidelines}?
    \item[] Answer: \answerYes{}
    \item[] Justification: \answerNA{}

\item {\bf Broader Impacts}
    \item[] Question: Does the paper discuss both potential positive societal impacts and negative societal impacts of the work performed?
    \item[] Answer: \answerYes{} 
    \item[] Justification: \answerNA{}
    
\item {\bf Safeguards}
    \item[] Question: Does the paper describe safeguards that have been put in place for responsible release of data or models that have a high risk for misuse (e.g., pretrained language models, image generators, or scraped datasets)?
    \item[] Answer: \answerNA{} 
    \item[] Justification: \answerNA{}

\item {\bf Licenses for existing assets}
    \item[] Question: Are the creators or original owners of assets (e.g., code, data, models), used in the paper, properly credited and are the license and terms of use explicitly mentioned and properly respected?
    \item[] Answer: \answerYes{} 
    \item[] Justification:  \answerNA{}.

\item {\bf New Assets}
    \item[] Question: Are new assets introduced in the paper well documented and is the documentation provided alongside the assets?
    \item[] Answer: \answerNA{} 
    \item[] Justification: \answerNA{}

\item {\bf Crowdsourcing and Research with Human Subjects}
    \item[] Question: For crowdsourcing experiments and research with human subjects, does the paper include the full text of instructions given to participants and screenshots, if applicable, as well as details about compensation (if any)? 
    \item[] Answer: \answerNA{} 
    \item[] Justification: \answerNA{}

\item {\bf Institutional Review Board (IRB) Approvals or Equivalent for Research with Human Subjects}
    \item[] Question: Does the paper describe potential risks incurred by study participants, whether such risks were disclosed to the subjects, and whether Institutional Review Board (IRB) approvals (or an equivalent approval/review based on the requirements of your country or institution) were obtained?
    \item[] Answer: \answerNA{}
    \item[] Justification: \answerNA{}

\end{enumerate}